\documentclass[journal]{IEEEtran}

\usepackage{amsmath,amssymb}
\usepackage{graphicx}
\usepackage{booktabs}
\usepackage{multirow}
\usepackage{array}
\usepackage{makecell}
\usepackage{cite}
\usepackage{url}
\usepackage{xcolor}
\usepackage{stfloats}
\usepackage{balance}

\newcommand{\method}{\textsc{PredRA}}
\newcommand{\methodddpm}{\textsc{PredRA-DDPM}}
\newcommand{\methodflow}{\textsc{PredRA-Flow}}
\newcommand{\tablefont}{\scriptsize}
\newcommand{\best}[1]{\textbf{#1}}
\newcommand{\secondbest}[1]{\underline{#1}}

\begin{document}

\title{PredRA: Fast Medical Image Translation by Deterministic Component Extraction and Controlled Stochastic Refinement}

\author{Jianhai Zhang$^{1,2,*}$, Pattarawut Charatpangoon$^{1,2}$, Donghao Zhang$^{3}$, Bijoy K. Menon$^{1,2}$, Wu Qiu$^{3}$,\\ M. Ethan MacDonald$^{2,4}$, and Aravind Ganesh$^{1,2,*}$%
\thanks{$^{1}$Department of Clinical Neurosciences, Cumming School of Medicine, University of Calgary, Calgary, AB, Canada.}%
\thanks{$^{2}$Ron and Rene Ward Centre for Healthy Brain Aging Research, Hotchkiss Brain Institute, University of Calgary, Calgary, AB, Canada.}%
\thanks{$^{3}$Department of Biomedical Engineering, School of Life Science and Technology, Huazhong University of Science and Technology, Wuhan, Hubei, China.}%
\thanks{$^{4}$Department of Biomedical Engineering, Schulich School of Engineering, University of Calgary, Calgary, AB, Canada.}%
\thanks{$^{*}$Corresponding authors: Jianhai Zhang (jianhai.zhang@ucalgary.ca) and Aravind Ganesh (aganesh@ucalgary.ca).}}

\markboth{}%
{Zhang \MakeLowercase{\textit{et al.}}: PredRA for Fast Medical Image Translation}

\maketitle

\begin{abstract}
Strongly paired medical image translation contains a substantial component that can be predicted directly from the source. The current reality is that pure deterministic prediction can smooth away fine detail, while generative models can recover detail but may also introduce unnecessary or potentially harmful variation. We propose PredRA\kern-0.6em\begingroup
\renewcommand{\thefootnote}{\ensuremath{\dagger}}
\footnote{Code is available at \url{https://github.com/jianhai-zhang/RredRA}.}
\endgroup\kern-0.6em, a fast framework that uses the deterministic prediction as a stable reference and further extracts additional deterministic components from the residual during generative refinement, thereby improving fidelity while maintaining perceptual quality.
The goal is simple: we view the entire residual-based generative process as an optimization problem and derive a practical solution that recovers useful residual detail through controlled refinement while keeping the prediction close to the paired target. Mechanism studies further show that useful residual information follows structured patterns, but its usefulness is difficult to estimate reliably at the voxel level. We therefore globally control how much of the residual refinement is added to the deterministic prediction, thereby reducing the accumulation of unnecessary uncertainty.
PredRA therefore combines deterministic component extraction with controlled stochastic refinement for fast, fidelity-preserving, and perceptually strong medical image translation. We validate the approach across multiple real-world datasets, showing that controlled refinement improves fidelity to the paired target compared with full residual refinement while retaining much of the perceptual benefit of generative modeling. PredRA achieves competitive or superior performance to substantially larger state-of-the-art models with 1.4--11.9$\times$ fewer total parameters and 3.1--26.2$\times$ fewer trainable parameters, while its 32-step flow sampler requires 31.25$\times$ fewer sampling steps than the matched 1000-step DDPM.
\end{abstract}


\section{Introduction}
\IEEEPARstart{M}{edical} image translation predicts target images from available source images. Existing approaches include conditional GANs, transformer/state-space models, diffusion models, and flow-based generators~\cite{isola2017pix2pix,zhu2017cyclegan,dalmaz2022resvit,atli2026i2imamba,ho2020ddpm,kazerouni2023survey}. These methods can produce realistic images, but strongly paired medical translation has an important property: \textit{much of the target is already determined by the source and should not be changed unnecessarily.}

This creates a natural trade-off between fidelity and perceptual quality. Deterministic models usually preserve paired structure well, but they can smooth fine detail. Generative models can recover detail, but stronger generation can also introduce variation that is not needed for the paired target. The useful solution is therefore not to choose only deterministic or only generative inference. Unlike existing coarse-to-fine generative pipelines, we explicitly treat the deterministic prediction as the stable reference and use generation only for the residual information that remains unexplained, adding back only the amount of refinement supported by validation.

The remaining residual is important because it
can still contain an additional predictable component. To extract it from the residual, we write this residual as a stable source-supported part plus a remaining zero-mean term. A residual generator is then used to model the full correction. In practice, however, separating the useful stable part from sampling-dependent variation at every voxel is difficult. Our main idea is therefore to control the residual globally: add enough generated correction to recover useful detail, but stop before unnecessary generative variation begins to reduce paired fidelity.

Specifically, a deterministic network first produces a stable reference $\hat Y_{\mathrm{det}}$. A residual-based generative model then generates one correction proposal $\tilde r_{\phi}$. Instead of always accepting the full correction, we form
\begin{equation}
\hat Y_{\lambda}=\hat Y_{\mathrm{det}}+\lambda\tilde r_{\phi},
\end{equation}
and select optimal $\lambda^\star$ on validation data. The entire refinement process is treated as a simple constrained optimization problem: maximize paired fidelity while keeping the perceptual benefit of generation within a fixed tolerance. The same realized residual proposal is reused across candidate $\lambda$ values, so this comparison changes only how much refinement is accepted.

We evaluate the idea on multiple real-world medical imaging datasets. For each task, we compare the deterministic reference, the full DDPM/Flow residual proposal, the validation-selected \method{} result, and published translation baselines. The matched Full-versus-PredRA comparison directly tests whether controlled use of the same residual proposal improves the final prediction.

The main contributions are straightforward. First, we keep a deterministic prediction as a stable reference rather than allowing generation to replace it. Second, we show that the residual still contains useful structured information after deterministic prediction, and that partial refinement can recover this information while avoiding unnecessary or potentially harmful sampling-dependent variation. Third, we formulate the final residual strength as a validation-guided optimization that favors fidelity while preserving perceptual quality. Fourth, the same idea works with conditional DDPM and  Flow, and Flow provides the fast operating variant. Finally, spatial and trajectory studies show why a simple global controller is more reliable than a learned voxel-wise gate in the current experiments.

\section{Related Work}
\subsection{Deterministic, Adversarial, and State-Space Translation}
Paired conditional GANs such as Pix2Pix established a direct source-to-target mapping paradigm~\cite{isola2017pix2pix}, while CycleGAN enabled unpaired translation through cycle consistency~\cite{zhu2017cyclegan}. Medical variants have emphasized anatomical fidelity, modality handling, and long-range context. MedGAN combined adversarial and non-adversarial objectives for end-to-end medical image translation~\cite{armanious2020medgan}; ResViT combined convolutional and transformer components for multimodal synthesis~\cite{dalmaz2022resvit}; unified missing-modality synthesis supports different combinations of available inputs within a single model~\cite{zhang2025unified}; and I2I-Mamba uses selective state-space modeling to capture long-range context while avoiding the attention-complexity trade-offs emphasized by transformer-based designs~\cite{atli2026i2imamba}.

Earlier medical synthesis studies established paired and unpaired cross-modality translation with context-aware GANs, modality-invariant latent representations, conditional multi-contrast synthesis, cycle-consistent MR--CT translation, and anatomy- or segmentation-aware constraints~\cite{nie2017context,chartsias2018multimodal,dar2019multicontrast,wolterink2017unpaired,zhang2018shape,huo2019synseg}. General unsupervised image-to-image translation also developed shared-latent and multimodal formulations that separate domain-invariant content from domain-specific variation~\cite{liu2017unit,huang2018munit}.

These studies support a simple starting point: keep the deterministic prediction as a stable reference. PredRA then asks a more focused question: after this first prediction, how much useful structure is still left in the residual, and how much of a generated residual correction should be added back?

\subsection{Diffusion, Flow Matching, and Diffusion Bridges for Translation}
DDPMs and score-based models construct reverse stochastic processes from progressively perturbed data~\cite{ho2020ddpm,song2021sde}. Palette demonstrated a general conditional-diffusion framework for image-to-image translation~\cite{saharia2022palette}, while BBDM and BiBBDM use Brownian-bridge processes to connect source and target domains~\cite{li2023bbdm,xue2025bibbdm}. Flow Matching provides a simulation-free objective for learning continuous-normalizing-flow vector fields along prescribed probability paths and offers an ODE-based alternative to stochastic reverse diffusion~\cite{lipman2023flow}. We use conditional flow matching as a second residual generator to test whether the refinement principle depends on DDPM.

Medical imaging has rapidly adopted these generative formulations. Khader \emph{et al.} studied 3-D medical-image generation with diffusion models, Pinaya \emph{et al.} introduced latent diffusion for 3-D brain-image generation, and Medfusion compared conditional latent DDPMs with GANs across medical imaging domains~\cite{khader2023medical,pinaya2022brain,mullerfranzes2023medfusion}. For translation, SynDiff combines adversarial and diffusion components in an unpaired framework~\cite{ozbey2023syndiff}; FGDM performs zero-shot translation with frequency-guided diffusion~\cite{li2024fgdm}; mutual-information guidance provides another zero-shot cross-modality mechanism~\cite{wang2024midiffusion}; TGDM uses target-guided diffusion for unpaired cross-modality translation~\cite{luo2024tgdm}; and M2DN conditions diffusion on modality-availability masks for arbitrary missing-MRI synthesis~\cite{meng2024m2dn}. These studies show that source preservation and modality guidance are already central problems; our contribution is not a new conditioning mechanism.

\subsection{Deterministic Priors, Residual Diffusion, and Frequency Decomposition}
Several existing methods already use residual or prior-guided diffusion, so our novelty is not residual diffusion itself. ResShift constructs a residual-shifting diffusion process for restoration~\cite{yue2023resshift}. RDDM explicitly decouples residual diffusion from noise diffusion~\cite{liu2024rddm}, and DRDD further separates stochastic noise diffusion from a deterministic residual-diffusion stage for image-to-image translation~\cite{lin2026drdd}. MRDPM is a residual-diffusion method for perfusion synthesis~\cite{cai2024mrdpm}. Thus, residual diffusion is prior art rather than the novelty of this paper.

Likewise, using a deterministic image before or inside diffusion is established. CMDM first obtains a conditional-GAN prior and then applies cascaded multi-path shortcut diffusion, with multiple reverse paths also used for uncertainty estimation~\cite{zhou2024cmdm}. MADM uses a CNN-generated 3-D prior to accelerate a 2.5-D multi-view averaging diffusion model~\cite{chen2025madm}. SelfRDB defines a source-to-target diffusion bridge and recursively refines transient target estimates toward a self-consistent solution~\cite{arslan2025selfrdb}. FDDM first estimates target-domain anatomical information and then applies a frequency-decoupled dual-path diffusion stage~\cite{li2025fddm}.  

\subsection{Fidelity, Uncertainty, and Controlled Stochasticity}
The most direct conceptual prior is the perception--distortion trade-off~\cite{blau2018tradeoff}. Rassmann \emph{et al.} recently showed in medical image translation that iterative generative sampling can improve perceptual appearance while reducing information fidelity, and studied regression/expectation-approximation alternatives~\cite{rassmann2026regression}. Our study uses the middle ground: keep the deterministic prediction, generate one residual correction, and choose how much of that correction to add back.

Trajectory reliability is also an established diffusion-model concern. Residual-learning work has attributed final-image error to accumulated score-estimation and discretization errors and learned a correction function for the reverse sampling trajectory~\cite{zhang2024trajres}. The oracle-to-free trajectory analysis therefore serves as complementary mechanism evidence for paired translation.

Uncertainty and spatial adaptation have also been incorporated into diffusion in several distinct ways. MGDM introduces uncertainty guidance during sampling~\cite{luo2024mgdm}; UPSR uses an uncertainty estimate to modulate region-specific perturbation strength~\cite{zhang2025upsr}; CMDM estimates translation uncertainty from multiple shortcut paths~\cite{zhou2024cmdm}; and conformal risk-control methods provide calibrated risk guarantees for diffusion outputs~\cite{teneggi2023trust}. More broadly, aleatoric and epistemic uncertainty are distinct concepts in predictive deep learning~\cite{kendall2017uncertainty}. Our learned spatial-gating experiment is therefore a negative ablation rather than a first uncertainty-guided method. Here, ``uncertainty'' is used in a practical structural sense. The deterministic model removes as much paired error as it can; the residual generator handles what remains; and the residual scale limits how much sampling-dependent variation is added.

\begin{figure*}[t]
\centering
\includegraphics[width=\textwidth]{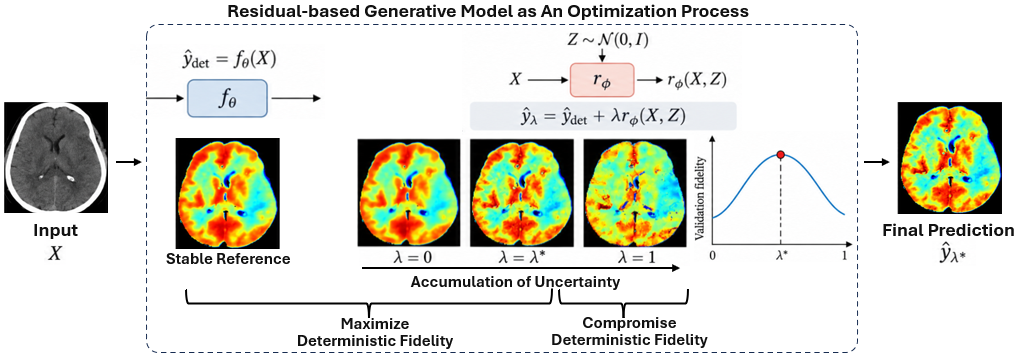}
\caption{Overview of \method{}. The deterministic model first captures the part of the target that can be predicted directly from the source. The residual generator then models what remains, including useful stable residual structure that the first model missed. \method{} reuses one residual proposal and varies only its strength $\lambda$. Validation selects $\lambda^*$ to maximize SSIM while keeping LPIPS within the fixed constraint in~\eqref{eq:constrainedlambda}. In this way, the final prediction recovers additional residual detail without accepting unnecessary generative variation. The schematic shows one CT input for clarity; the experiments also include CT+CTA inputs.}
\label{fig:mathframework}
\end{figure*}

\section{Deterministic Component Extraction and Residual Structure}

PredRA starts with a deterministic prediction because it gives the most direct estimate of the target information already supported by the source. We do not assume that this prediction is perfect or equal to the true conditional mean. We only use it as the strongest validation-supported deterministic reference under the fixed protocol.
Let $(X,Y)\sim p^{*}(x,y)$ denote a paired source--target sample. For a deterministic hypothesis class $\mathcal{F}$ and a prespecified paired fidelity loss $\ell_{\mathrm{det}}$, the population risk is
\begin{equation}
\mathcal{R}_{\mathrm{det}}(f)=
\mathbb{E}_{p^{*}(x,y)}\!\left[\ell_{\mathrm{det}}\!\left(Y,f(X)\right)\right].
\label{eq:detrisk}
\end{equation}
The best predictor within the chosen deterministic model class is
\begin{equation}
f^{*}_{\mathcal{F}}=\arg\min_{f\in\mathcal{F}}\mathcal{R}_{\mathrm{det}}(f),
\qquad
\hat Y_{\mathrm{det}}=f_{\theta}(X)\approx f^{*}_{\mathcal{F}}(X).
\label{eq:opdet}
\end{equation}
For squared error and an unrestricted function class, the Bayes solution is the conditional mean; Appendix~\ref{app:mathdetails} gives the proof. In practice, the network and training objective are finite, so $\hat Y_{\mathrm{det}}$ is simply the best deterministic reference obtained with the frozen protocol.

We use the remaining deterministic risk as a simple measure of how much paired error is still left before generation,
\begin{equation}
\mathcal{U}_{\mathrm{pre}}(\theta)
\equiv
\mathcal{R}_{\mathrm{det}}(f_{\theta}).
\label{eq:upre}
\end{equation}
This is not a calibrated posterior variance. It only measures how much paired target error remains before the generative model is used. The deterministic model therefore tries to make $\mathcal{U}_{\mathrm{pre}}$ as small as possible under the fixed protocol.

The paired residual is
\begin{equation}
R_{\theta}=Y-f_{\theta}(X).
\label{eq:rtheta}
\end{equation}
A useful conditional decomposition is
\begin{equation}
\begin{aligned}
R_{\theta}&=b_{\theta}(X)+\varepsilon_{\theta},\\
b_{\theta}(X)&=\mathbb{E}[R_{\theta}\mid X],
\qquad
\mathbb{E}[\varepsilon_{\theta}\mid X]=0.
\end{aligned}
\label{eq:resdecomp}
\end{equation}
The term $b_{\theta}(X)$ is the key part for our interpretation: it is additional source-supported structure that remains predictable because the first deterministic model is imperfect. We refer to it as the stable residual component. The remaining term $\varepsilon_{\theta}$ collects what is left after conditioning on the source; it is not assumed to be pure acquisition noise or pure aleatoric uncertainty.

This decomposition explains the main idea of PredRA. The residual can still contain useful, source-supported structure, but it can also contain sampling-dependent variation. We therefore test how predictable and spatially structured the residual is. These analyses help motivate the refinement controller, but they do not explicitly separate irreducible aleatoric uncertainty from other sources of variation.

\section{PredRA: Recovering Stable Residual Detail with Controlled Refinement}
\subsection{Residual Refinement as an Optimization Problem}
After the deterministic reference is frozen, the residual generator uses a random variable $Z$ to produce one correction proposal $\tilde r_{\phi}(X,Z)$. This proposal can contain both useful stable residual structure and sampling-dependent variation. Before projection to the valid output range, the prediction is
\begin{equation}
\bar Y_{\lambda}(X,Z)
=
\hat Y_{\mathrm{det}}(X)+\lambda\tilde r_{\phi}(X,Z),
\qquad \lambda\in[0,1].
\label{eq:unproj_random}
\end{equation}
The sampling variability in this prediction is
\begin{equation}
\mathcal{U}_{\mathrm{post}}(\lambda)
=
\mathbb{E}_{X}\!\left[
\operatorname{tr}\operatorname{Cov}_{Z}
\!\left(\bar Y_{\lambda}(X,Z)\mid X\right)
\right].
\label{eq:upost}
\end{equation}
Because the deterministic anchor is fixed with respect to $Z$,
\begin{equation}
\mathcal{U}_{\mathrm{post}}(\lambda)
=
\lambda^2\,
\mathbb{E}_{X}\!\left[
\operatorname{tr}\operatorname{Cov}_{Z}
\!\left(\tilde r_{\phi}(X,Z)\mid X\right)
\right].
\label{eq:upost_scale}
\end{equation}
Thus $\lambda$ has a simple meaning. It controls how much of the generated residual correction is accepted. Because the same residual realization is scaled, it also directly scales the amount of sampling variability added before projection. This does not claim calibrated predictive uncertainty. Appendix~\ref{app:mathdetails} gives the short derivation and discusses projection.

PredRA is trained in stages. The deterministic model first captures the easy-to-predict part. The residual model then learns the remaining correction, including stable residual structure that the first model missed. The final controller decides how much of this correction should be accepted by solving
\begin{equation}
\lambda^{*}
=
\arg\max_{\lambda\in\Lambda}
\mathcal{F}_{\mathrm{val}}(\lambda)
\quad\text{s.t.}\quad
\mathcal{C}_{\mathrm{gen}}(\lambda)\leq\tau,
\label{eq:unified_objective}
\end{equation}
where $\mathcal{F}_{\mathrm{val}}$ measures validation fidelity and $\mathcal{C}_{\mathrm{gen}}$ limits the loss of perceptual benefit from generation. In our experiments, $\mathcal{F}_{\mathrm{val}}$ is SSIM and $\mathcal{C}_{\mathrm{gen}}(\lambda)=P_{\mathrm{val}}(\lambda)-P_{\mathrm{val}}(1)$ with LPIPS $P_{\mathrm{val}}$. In short, we keep as much useful residual detail as possible while preventing unnecessary generative variation from moving the result away from the paired target.

\subsection{Conditional Residual DDPM}
Once the deterministic reference is fixed, the generator only needs to model what remains in the residual. Define the residual target
\begin{equation}
r^{\star}=Y-\hat Y_{\mathrm{det}}.
\end{equation}
For a variance schedule $\{\beta_t\}_{t=1}^{T}$, let $a_t=1-\beta_t$ and $\bar a_t=\prod_{s=1}^{t}a_s$. The forward perturbation at timestep $t$ is
\begin{equation}
q(r_t\mid r^{\star})=
\mathcal{N}\!\left(\sqrt{\bar a_t}\,r^{\star},(1-\bar a_t)I\right),
\end{equation}
or equivalently
\begin{equation}
r_t=\sqrt{\bar a_t}\,r^{\star}+\sqrt{1-\bar a_t}\,\epsilon,
\qquad \epsilon\sim\mathcal{N}(0,I).
\end{equation}
With conditioning $C=(X,\hat Y_{\mathrm{det}})$, the noise predictor is trained with
\begin{equation}
\mathcal{L}_{\mathrm{DDPM}}=
\mathbb{E}_{r^{\star},t,\epsilon}
\!\left[\left\|\epsilon-\epsilon_{\phi}(r_t,t,C)\right\|_2^2\right].
\end{equation}
At inference, the original ancestral DDPM reverse process is initialized from Gaussian noise and iterated to obtain one realized residual proposal $\tilde r_{\phi}$.

\subsection{Conditional Residual Flow Matching}
To test whether the same idea works beyond DDPM, we keep the same deterministic reference and residual target $r^{\star}$ and replace the DDPM reverse chain with conditional flow matching~\cite{lipman2023flow}. Draw $z\sim\mathcal{N}(0,I)$ and define the straight conditional path
\begin{equation}
r_t=(1-t)z+t r^{\star},\qquad t\in[0,1].
\end{equation}
For each sampled pair $(z,r^{\star})$, the path derivative is constant,
\begin{equation}
u_t=\frac{\mathrm{d}r_t}{\mathrm{d}t}=r^{\star}-z.
\end{equation}
A conditional vector field $v_{\phi}(r_t,t,C)$ is therefore trained by
\begin{equation}
\mathcal{L}_{\mathrm{Flow}}=
\mathbb{E}_{r^{\star},z,t}
\!\left[\left\|v_{\phi}(r_t,t,C)-(r^{\star}-z)\right\|_2^2\right].
\end{equation}
At inference, a fresh $z\sim\mathcal{N}(0,I)$ initializes $r(0)=z$, and the ODE $\mathrm{d}r/\mathrm{d}t=v_{\phi}(r,t,C)$ is integrated to $t=1$. The implemented evaluation uses explicit Euler integration and evaluates 1, 2, 4, 8, 16, 32, and 64 steps. The DDPM and Flow models use the same residual U-Net. Matched-width comparisons therefore change the training/sampling method, not the residual-network family.

\subsection{Validation-Guided Refinement Strength}
For a fixed realized residual proposal $\tilde r_{\phi}$ from either backend, the test-time output is
\begin{equation}
\hat Y_{\lambda}=
\Pi_{\mathcal{Y}}\!\left(\hat Y_{\mathrm{det}}+\lambda\tilde r_{\phi}\right),
\qquad \lambda\in[0,1],
\label{eq:lambda}
\end{equation}
where $\Pi_{\mathcal{Y}}$ projects the result to the fixed target range for each task. Here $\lambda=0$ gives the deterministic reference and $\lambda=1$ accepts the full residual proposal. The same generated residual is reused for every candidate $\lambda$, so the only change is how much refinement is accepted. Appendix~\ref{app:mathdetails} shows why an intermediate value $0<\lambda<1$ can be optimal when the residual proposal is imperfect.

Equation~\eqref{eq:unified_objective} is instantiated with validation SSIM as the fidelity objective and LPIPS retention as the generative constraint. Let $S_{\mathrm{val}}(\lambda)$ denote validation SSIM (higher is better) and $P_{\mathrm{val}}(\lambda)$ validation LPIPS (lower is better). The frozen controller selects
\begin{equation}
\lambda^{*}=
\arg\max_{\lambda\in\Lambda}S_{\mathrm{val}}(\lambda)
\quad
\text{s.t.}\quad
P_{\mathrm{val}}(\lambda)\leq P_{\mathrm{val}}(1)+\delta,
\label{eq:constrainedlambda}
\end{equation}
with the prespecified LPIPS slack $\delta=0.0050$. PSNR and MAE are used only as fixed tie-breakers. This is a simple validation-based selection rule, not probabilistic calibration. It chooses the setting with the best paired fidelity while requiring it to keep the LPIPS benefit of the full residual proposal within the prespecified slack. Throughout the paper, ``Full-DDPM'' and ``Full-Flow'' denote the corresponding full-strength $\lambda=1$ residual proposals rather than separately trained direct target generators.

\section{Experimental Design}
\subsection{Translation Tasks}
We evaluate three translation settings. The primary CT+CTA$\rightarrow$CTP cohort predicts CBF, CBV, and Tmax and uses a frozen patient-level split of 55 training, 12 validation, and 12 test patients. A second registered stroke-imaging experiment predicts joint DWI and ADC from CT and CTA using a frozen 70/10/20 patient split, all frozen test slices, the same canonical $192\times192$ field, and slice-to-patient-to-cohort aggregation. For this task, CT, CTA, DWI, and ADC each use one train-only robust affine scaler that is frozen for validation and test rather than per-case normalization.

The third setting uses 180 paired CT/MR cases from the SynthRAD brain cohort, split into 126 training, 27 validation, and 27 frozen test cases. CT is the source and MR is the target; this is an experimental reverse-direction use of the paired SynthRAD data rather than the official benchmark direction. CT is windowed to $[0,100]$, MR is normalized within the intracranial support using robust per-volume percentiles, and final evaluation uses the complete frozen test manifest on the canonical $192\times192$ field.

\subsection{Comparators and Matched Controls}
The mechanism comparison includes the deterministic reference ($\lambda=0$), the full residual proposal ($\lambda=1$), \method{} with validation-selected residual strength, a ground-truth spatial oracle, and the best GT-free spatial gate. The CT+CTA$\rightarrow$CTP benchmark additionally places SelfRDB~\cite{arslan2025selfrdb}, I2I-Mamba~\cite{atli2026i2imamba}, ResViT~\cite{dalmaz2022resvit}, MADM~\cite{chen2025madm}, and YODA~\cite{rassmann2026regression} under the same frozen input cache, test slices, canonical $192\times192$ field, and metric implementation. Multi-input/multi-output interface adaptations are used where required, but the native backbone family and training objective of each comparator are retained. The CTP allocation controller is target specific, so the table reports it as validation-selected rather than forcing three target-wise strengths into one scalar column. The CT+CTA$\rightarrow$DWI+ADC experiment reuses this all-method interface and evaluator template. The matched DET, Full-DDPM/Full-Flow, and PredRA-DDPM/PredRA-Flow comparisons remain the primary matched test of controlled residual refinement on this task.

The independent SynthRAD benchmark further includes official reproductions of I2I-Mamba~\cite{atli2026i2imamba} and ResViT~\cite{dalmaz2022resvit}, plus protocol-adapted reproductions of MADM~\cite{chen2025madm} and YODA~\cite{rassmann2026regression}. MADM requires a precomputed CNN prior, but the public repository does not release prior-generator training code; we therefore use a train-only 2-D U-Net prior selected by validation patient MAE and disclose this as the only substantive MADM protocol adaptation. YODA is evaluated in reverse CT$\rightarrow$MRI direction, which is outside the public model zoo directions; its final checkpoint and sampling mode are selected on validation only. YODA additionally required an explicit unit-range adapter because the official loader assumes 8-bit 0--255 input, whereas the frozen shared SynthRAD export is float32 in 0--1. The corrected adapter maps the frozen data exactly into YODA's $[-1,1]$ model domain, decodes official uint8 predictions by fixed division by 255, and restores predictions to the canonical grid using the saved NIfTI affine.

All published comparators use validation-only checkpoint/model selection. The final SynthRAD comparison is then recomputed under one strict frozen-test contract: all 27 test patients, every row in the frozen test manifest, the canonical $192\times192$ field after undoing any interface-only padding or canvas expansion, one shared target/mask source, and one shared metric implementation. No fixed-$k$ slice subset is used in the final SynthRAD table. Coverage is exact at the $(\text{case ID},z)$ level; a method missing any required test slice is treated as incomplete rather than mixed into the comparison. Test metrics are not used to choose checkpoints, residual strength, preprocessing, or comparator configuration for the primary frozen comparisons. The later residual-capacity and Flow step-count sweeps are explicitly treated as post hoc sensitivity analyses because they were extended after frozen-test inspection.

The deterministic capacity study is performed on the full validation set and is used to select the 15.254M-parameter DET-15.25M anchor before the residual-backend retraining experiments. With DET-15.25M fixed, the residual-network capacity is then varied over 1.547M, 3.310M, 5.745M, 8.851M, and 12.628M parameters for both DDPM and Flow. The U-Net width hyperparameter $b$ denotes the number of channels in the first feature stage; because the same $b$ maps to different parameter counts in the deterministic and residual architectures, reader-facing model names use actual parameter counts instead. Capacity labels round parameter counts to two decimals (for example, 15.254M$\rightarrow$DET-15.25M and 12.628M$\rightarrow$12.63M), while exact counts are retained in the capacity and efficiency tables. For Full/PredRA names, the capacity suffix denotes the residual-network size rather than the total two-stage system size; DET-15.25M plus the 12.628M residual model contains 27.882M unique parameters. Each residual capacity receives its own validation-selected strength $\lambda$. Because the backend-capacity sweep was extended after frozen-test inspection, its test-set curves are reported only as post hoc sensitivity analyses and the 12.63M point is not presented as a validation-selected capacity. Flow sampling is likewise evaluated at 1, 2, 4, 8, 16, 32, and 64 integration steps as a post hoc sensitivity analysis. The matched Full-versus-PredRA comparison at any fixed architecture remains the primary controlled test of residual refinement.

\subsection{Mechanistic Validation Protocols}
The spatial-control analysis generalizes the scalar controller to a map $A:\Omega\rightarrow[0,1]$,
\begin{equation}
\hat Y_A=\Pi_{\mathcal{Y}}\!\left(\hat Y_{\mathrm{det}}+A\odot\tilde r_{\phi}\right).
\end{equation}
A ground-truth-informed oracle asks whether spatially varying residual utility exists at all; a test-time gate must infer $A$ without target information, using only source images, the deterministic prediction, the realized residual, and optional trajectory features. This separates the existence of spatial heterogeneity from its deployable predictability.

For the reverse-trajectory analysis, let $r_t^{\mathrm{or}}$ be a target-consistent state produced by the known forward process and $r_t^{\mathrm{fr}}$ the corresponding state reached by autonomous reverse sampling. Using a higher-is-better score $S$ (SSIM in this analysis), intervention $M$, and baseline $B$, define
\begin{align}
\Delta_t^{\mathrm{or}} &=
S(\hat Y_t^{M}(r_t^{\mathrm{or}}),Y)
-S(\hat Y_t^{B}(r_t^{\mathrm{or}}),Y),\label{eq:oracle_delta}\\
\Delta_t^{\mathrm{fr}} &=
S(\hat Y_t^{M}(r_t^{\mathrm{fr}}),Y)
-S(\hat Y_t^{B}(r_t^{\mathrm{fr}}),Y).\label{eq:free_delta}
\end{align}
An oracle-to-free reversal is recorded when $\Delta_t^{\mathrm{or}}>0$ but $\Delta_t^{\mathrm{fr}}\leq0$. The analysis is a mechanism test only: final model decisions are based on autonomous outputs and validation-level allocation, not on target-consistent intermediate states that are unavailable at deployment.

\subsection{Metrics and Aggregation}
Primary paired metrics are MAE, SSIM~\cite{wang2004ssim}, and PSNR. LPIPS~\cite{zhang2018lpips} is computed in the original \emph{unmasked} manner after preprocessing; no support-bounding-box or target-mask multiplication is applied inside LPIPS. For the final SynthRAD comparison, MAE and PSNR use the common intracranial support, SSIM uses the full skull-stripped canonical field, and LPIPS receives the same skull-stripped canonical prediction/target without an additional metric mask. Every slice is scored first, then averaged within patient, and finally averaged across the 27 frozen test patients.

External architectures may require tensor sizes different from the frozen project field. Such resizing or padding is treated only as model-interface preprocessing; all metrics are computed after restoring predictions to the canonical $192\times192$ evaluation field. A matched I2I-Mamba evaluation showed why this is necessary: its $256\times256$ interface canvas contained 43.7500\% zero padding, which artificially increased full-image SSIM. Cropping back to the canonical field left masked MAE and PSNR unchanged but changed SSIM by $-0.1726$ and LPIPS by $+0.0553$. The same canonical-domain rule is applied to every comparator, including the protocol-adapted MADM prior. These diagnostic values are not mixed with the final all-slice benchmark.

\section{Results}
\subsection{Stable Structure Remaining in the Residual}
In the CTP development experiments, held-out residual magnitude showed moderate predictability (Spearman $\rho\approx+0.4700$). This supports the idea that the residual still contains structured, source-related information after deterministic prediction. However, the signal is not strong enough to support a reliable voxel-wise controller. We therefore keep one global validation-selected refinement strength. This result does not mean that the residual is pure aleatoric uncertainty.

\subsection{Local Denoising Improvements Do Not Necessarily Survive Autonomous Sampling}
The oracle-to-free analysis directly evaluated the quantities defined in Eqs.~\eqref{eq:oracle_delta}--\eqref{eq:free_delta}. Improvements observed from target-consistent intermediate states did not always persist when the same comparison was made on states generated by the model itself. At the strongest representative high-noise reversal, the intervention improved SSIM by $+0.0144$ on the target-consistent state but changed SSIM by $-0.0003$ on the autonomous reverse trajectory. Additional reversals were observed at other starting timesteps, including cases in which a small positive oracle-state gain became negative after free-running evolution. Additional timestep-level analysis details are reported in Appendix~\ref{app:mechanism}.

The key result is the sign reversal: better denoising on an oracle-like intermediate state does not guarantee a better final paired image. We therefore judge residual changes using the actual free-running trajectory and the final validation criterion, not oracle intermediate states.

\subsection{Frozen-Test Results Across Tasks}
The CT+CTA$\rightarrow$CTP benchmark is now complete under a single validated evaluator. All methods use the same frozen patient split, all frozen test slices, the canonical $192\times192$ field, and the same slice-to-patient-to-cohort aggregation. This replaces the earlier heterogeneous CTP comparator values in the main benchmark while preserving the earlier mechanism analyses as development evidence.

The matched-backbone CTP comparisons directly support controlled residual refinement. Relative to Full-DDPM-12.63M, PredRA-DDPM-12.63M reduces MAE from 0.1899 to 0.1759, increases SSIM from 0.8053 to 0.8218, and increases PSNR from 12.9849 to 13.8127~dB, while LPIPS changes only from 0.1193 to 0.1210. Relative to Full-Flow32-12.63M, PredRA-Flow32-12.63M reduces MAE from 0.1720 to 0.1679, increases SSIM from 0.8117 to 0.8238, and increases PSNR from 13.7019 to 14.0907~dB, with LPIPS changing from 0.1140 to 0.1166. On the frozen test set these LPIPS increases are $+0.0018$ and $+0.0026$, respectively; they are descriptive test outcomes, whereas the $0.0050$ constraint in Eq.~\eqref{eq:constrainedlambda} is applied only on validation. All three paired fidelity metrics improve for both backends.

Across all methods, no single model is best on every metric. YODA gives the strongest CTP MAE, SSIM, and PSNR (0.1542/0.8521/14.9571~dB), whereas PredRA-Flow32-12.63M gives the lowest LPIPS among the ranked comparison methods (0.1166) and the second-best MAE and PSNR. The deterministic anchor has higher SSIM than both allocated generators but substantially higher LPIPS, while the full Flow proposal attains the lowest LPIPS overall at the cost of lower paired fidelity. MADM tracks the deterministic anchor closely in distortion metrics but remains higher in LPIPS than PredRA-Flow32-12.63M. These results do not show one universally best model. They show that PredRA improves its matched full residual proposal and gives a different fidelity--perception trade-off. Per-target CBF/CBV/Tmax values are reported in Appendix Table~\ref{tab:ctppertarget}.

The additional CT+CTA$\rightarrow$DWI+ADC benchmark gives an even stronger matched-backbone replication. Relative to Full-DDPM-12.63M, PredRA-DDPM-12.63M reduces macro MAE from 0.1951 to 0.1633, increases SSIM from 0.7776 to 0.8061, increases PSNR from 19.2924 to 21.2959~dB, and also lowers LPIPS from 0.1312 to 0.1283. Relative to Full-Flow32-12.63M, PredRA-Flow32-12.63M reduces MAE from 0.2103 to 0.1642, increases SSIM from 0.7501 to 0.8044, increases PSNR from 19.5377 to 21.4120~dB, and lowers LPIPS from 0.1480 to 0.1351. Thus, unlike the CTP benchmark where allocation trades a small LPIPS increase for substantially recovered paired fidelity, the DWI+ADC task improves all four macro metrics for both matched backends. Among the ranked comparison methods (excluding the DET and Full-* reference rows), YODA has the lowest macro MAE, PredRA-DDPM-12.63M has the highest SSIM and lowest LPIPS, and PredRA-Flow32-12.63M has the highest PSNR. This different ranking across metrics again shows that the methods make different fidelity--perception trade-offs. The full macro comparison is given in Table~\ref{tab:dwiadcmacro}, with DWI+ADC target-wise values in Appendix Table~\ref{tab:dwiadcpertarget}.

\begin{table}[t]
\caption{CT+CTA$\rightarrow$DWI+ADC frozen-test benchmark under the shared evaluator. \best{Bold} and \secondbest{underlining} mark the best and second-best among the ranked comparison methods; DET and Full-* rows are reference settings and are excluded from that ranking. External baselines use the compute-bounded schedules described in the text.}
\label{tab:dwiadcmacro}
\centering
\tablefont
\setlength{\tabcolsep}{1.35pt}
\begin{tabular}{@{}lrrrr@{}}
\toprule
Method & MAE$\downarrow$ & SSIM$\uparrow$ & PSNR$\uparrow$ & LPIPS$\downarrow$\\
\midrule
SelfRDB~\cite{arslan2025selfrdb} & 0.2636 & 0.7173 & 15.9121 & 0.2202\\
I2I-Mamba~\cite{atli2026i2imamba} & 0.2492 & 0.7229 & 15.8951 & 0.2379\\
ResViT~\cite{dalmaz2022resvit} & 0.2538 & 0.7139 & 15.8931 & 0.2700\\
MADM~\cite{chen2025madm} & 0.3171 & 0.7228 & 16.9847 & 0.2567\\
YODA~\cite{rassmann2026regression} & \best{0.1620} & \secondbest{0.8048} & 20.7667 & 0.1395\\
\cmidrule(l){1-5}
DET-15.25M & 0.1657 & 0.8033 & 21.2576 & 0.1389\\
Full-DDPM-12.63M & 0.1951 & 0.7776 & 19.2924 & 0.1312\\
Full-Flow32-12.63M & 0.2103 & 0.7501 & 19.5377 & 0.1480\\
\cmidrule(l){1-5}
PredRA-DDPM-12.63M & \secondbest{0.1633} & \best{0.8061} & \secondbest{21.2959} & \best{0.1283}\\
PredRA-Flow32-12.63M & 0.1642 & 0.8044 & \best{21.4120} & \secondbest{0.1351}\\
\bottomrule
\end{tabular}
\end{table}

On independent SynthRAD CT$\rightarrow$MRI, deterministic capacity scaling on validation identifies DET-15.25M as the best tested deterministic anchor before the 29.89M/61.00M capacity plateau. On the frozen test set, DET-15.25M obtains MAE/SSIM/PSNR/LPIPS of 0.1139/0.7510/17.3619~dB/0.1823. With this anchor fixed, increasing residual-backend width improves both \methodddpm{} and \methodflow{} overall, although the trend is not strictly monotonic at every intermediate width. At the largest evaluated 12.628M residual-backend point, validation selects $\lambda=0.4500$ for PredRA-DDPM-12.63M and $\lambda=0.5000$ for PredRA-Flow32-12.63M. Their frozen-test results are 0.1147/0.7490/17.3000~dB/0.1537 and 0.1144/0.7494/17.3293~dB/0.1537, respectively. Thus, at matched residual-network capacity, Flow is slightly better in all three paired fidelity metrics, while achieving the same LPIPS.. The difference between the two allocated outputs is small, but the unallocated proposals separate more clearly: Full-Flow32-12.63M reaches 0.1221/0.7204/16.5824~dB/0.1471, compared with 0.1248/0.7139/16.3686~dB/0.1481 for Full-DDPM-12.63M. Full-Flow32-12.63M therefore improves MAE by 0.0026, SSIM by 0.0065, PSNR by 0.2140~dB, and LPIPS by 0.0010 while using 32 rather than 1000 sampling steps.

The reported 12.63M SynthRAD PredRA-Flow32-12.63M point remains close to DET-15.25M in paired fidelity (MAE $+0.0005$, SSIM $-0.0015$, PSNR $-0.0326$~dB) while lowering LPIPS by 0.0287 (15.7000\%). DET-15.25M and the two Full-* rows are treated as intermediate references rather than ranked methods in Table~\ref{tab:final_synthrad}. Among the ranked SynthRAD comparison methods under the same canonical all-slice evaluation, PredRA-Flow32-12.63M improves all four reported metrics over SelfRDB, I2I-Mamba, and ResViT. MADM remains stronger in MAE/SSIM/PSNR, whereas PredRA-Flow32-12.63M has lower LPIPS. Again, the methods occupy different fidelity--perception trade-offs rather than showing one universal winner.

\begin{table}[t]
\caption{CT+CTA$\rightarrow$CTP frozen-test macro benchmark. \best{Bold} and \secondbest{underlining} mark the best and second-best among the ranked comparison methods; DET and Full-* rows are reference settings and are excluded from that ranking. ``val.'' denotes target-specific validation-selected residual strengths.}
\label{tab:final}
\centering
\tablefont
\setlength{\tabcolsep}{1.15pt}
\begin{tabular}{@{}lcrrrr@{}}
\toprule
Method & $\lambda$ & MAE$\downarrow$ & SSIM$\uparrow$ & PSNR$\uparrow$ & LPIPS$\downarrow$\\
\midrule
SelfRDB~\cite{arslan2025selfrdb} & -- & 0.1929 & 0.8112 & 12.6603 & 0.1289\\
I2I-Mamba~\cite{atli2026i2imamba} & -- & 0.1888 & 0.8080 & 12.8245 & 0.1254\\
ResViT~\cite{dalmaz2022resvit} & -- & 0.1874 & 0.8125 & 12.8727 & \secondbest{0.1187}\\
MADM~\cite{chen2025madm} & -- & 0.1744 & \secondbest{0.8324} & 13.8996 & 0.1426\\
YODA~\cite{rassmann2026regression} & -- & \best{0.1542} & \best{0.8521} & \best{14.9571} & 0.1203\\
\cmidrule(l){1-6}
DET-15.25M & 0 & 0.1744 & 0.8347 & 13.9691 & 0.1627\\
Full-DDPM-12.63M & 1 & 0.1899 & 0.8053 & 12.9849 & 0.1193\\
Full-Flow32-12.63M & 1 & 0.1720 & 0.8117 & 13.7019 & 0.1140\\
\cmidrule(l){1-6}
PredRA-DDPM-12.63M & val. & 0.1759 & 0.8218 & 13.8127 & 0.1210\\
PredRA-Flow32-12.63M & val. & \secondbest{0.1679} & 0.8238 & \secondbest{14.0907} & \best{0.1166}\\
\bottomrule
\end{tabular}
\end{table}

\begin{table}[t]
\caption{SynthRAD CT$\rightarrow$MRI frozen-test benchmark under the common canonical all-slice evaluator. \best{Bold} and \secondbest{underlining} mark the best and second-best among the ranked comparison methods; DET and Full-* rows are reference settings and are excluded from that ranking.}
\label{tab:final_synthrad}
\centering
\tablefont
\setlength{\tabcolsep}{1.15pt}
\begin{tabular}{@{}lcrrrr@{}}
\toprule
Method & $\lambda$ & MAE$\downarrow$ & SSIM$\uparrow$ & PSNR$\uparrow$ & LPIPS$\downarrow$\\
\midrule
SelfRDB~\cite{arslan2025selfrdb} & -- & 0.1364 & 0.7038 & 15.7203 & 0.1900\\
I2I-Mamba~\cite{atli2026i2imamba} & -- & 0.1240 & 0.7140 & 16.4184 & 0.1563\\
ResViT~\cite{dalmaz2022resvit} & -- & 0.1184 & 0.7195 & 16.8226 & 0.1564\\
MADM~\cite{chen2025madm} & -- & \best{0.1079} & \best{0.7621} & \best{17.6970} & 0.1692\\
YODA~\cite{rassmann2026regression} & -- & 0.1539 & 0.6245 & 15.0081 & 0.3291\\
\cmidrule(l){1-6}
DET-15.25M & 0 & 0.1139 & 0.7510 & 17.3619 & 0.1823\\
Full-DDPM-12.63M & 1 & 0.1248 & 0.7139 & 16.3686 & 0.1481\\
Full-Flow32-12.63M & 1 & 0.1221 & 0.7204 & 16.5824 & 0.1471\\
\cmidrule(l){1-6}
PredRA-DDPM-12.63M & 0.4500 & 0.1147 & 0.7490 & 17.3000 & \secondbest{0.1537}\\
PredRA-Flow32-12.63M & 0.5000 & \secondbest{0.1144} & \secondbest{0.7494} & \secondbest{17.3293} & \best{0.1537}\\
\bottomrule
\end{tabular}
\end{table}

\begin{figure*}[t]
\centering
\includegraphics[width=\textwidth]{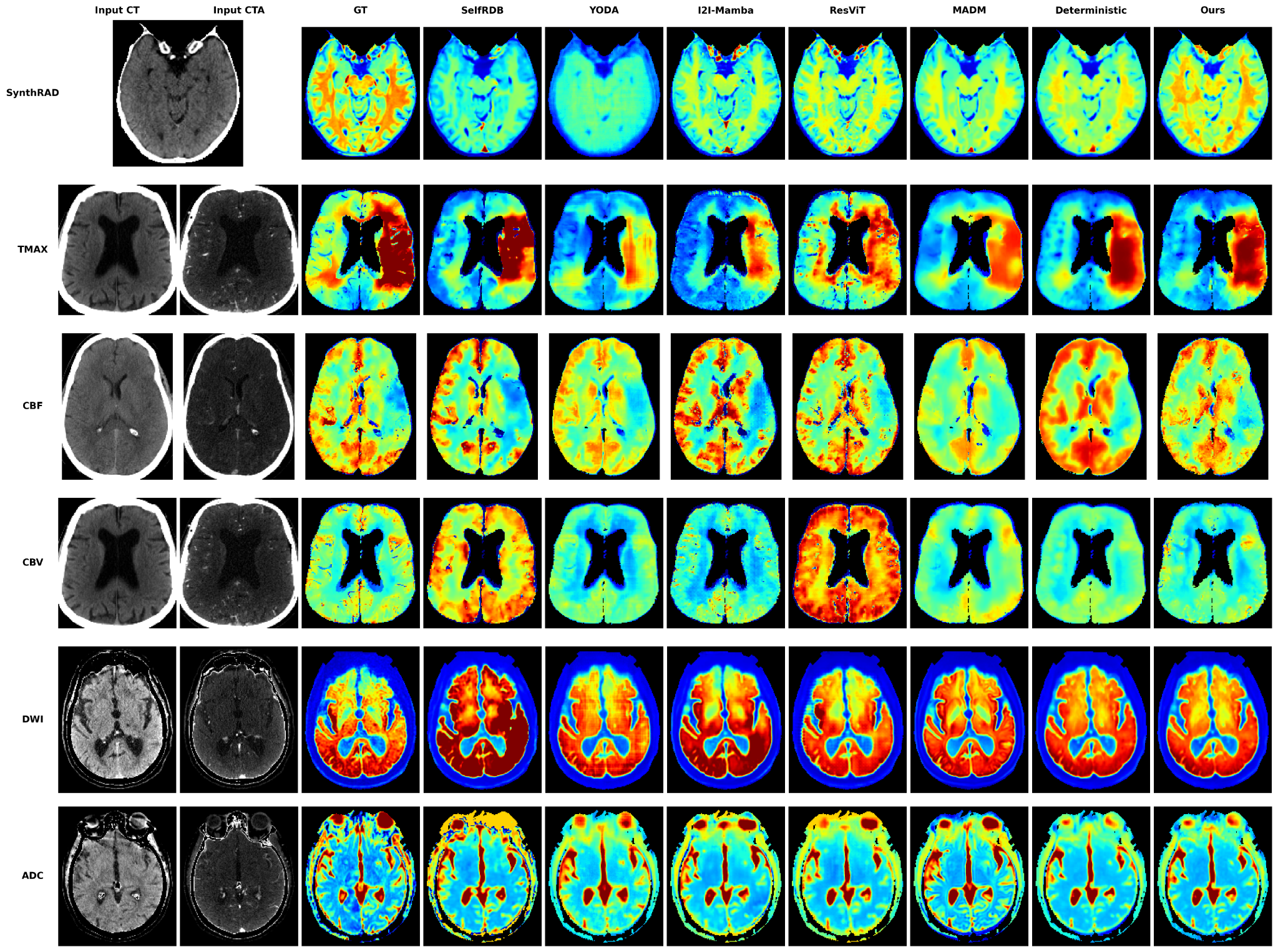}
\caption{Qualitative comparison across the three translation settings. Rows show SynthRAD CT$\rightarrow$MRI, CT+CTA$\rightarrow$CTP targets (Tmax, CBF, and CBV), and CT+CTA$\rightarrow$DWI+ADC. Columns compare the available input image(s), paired ground truth, five external baselines, the deterministic anchor, and the selected PredRA result. All examples use fixed frozen-test cases and the same display protocol within each target. Quantitative conclusions are based on the complete frozen test sets.}
\label{fig:qualitative_multitask}
\end{figure*}

\begin{figure*}[t]
\centering
\includegraphics[width=\textwidth]{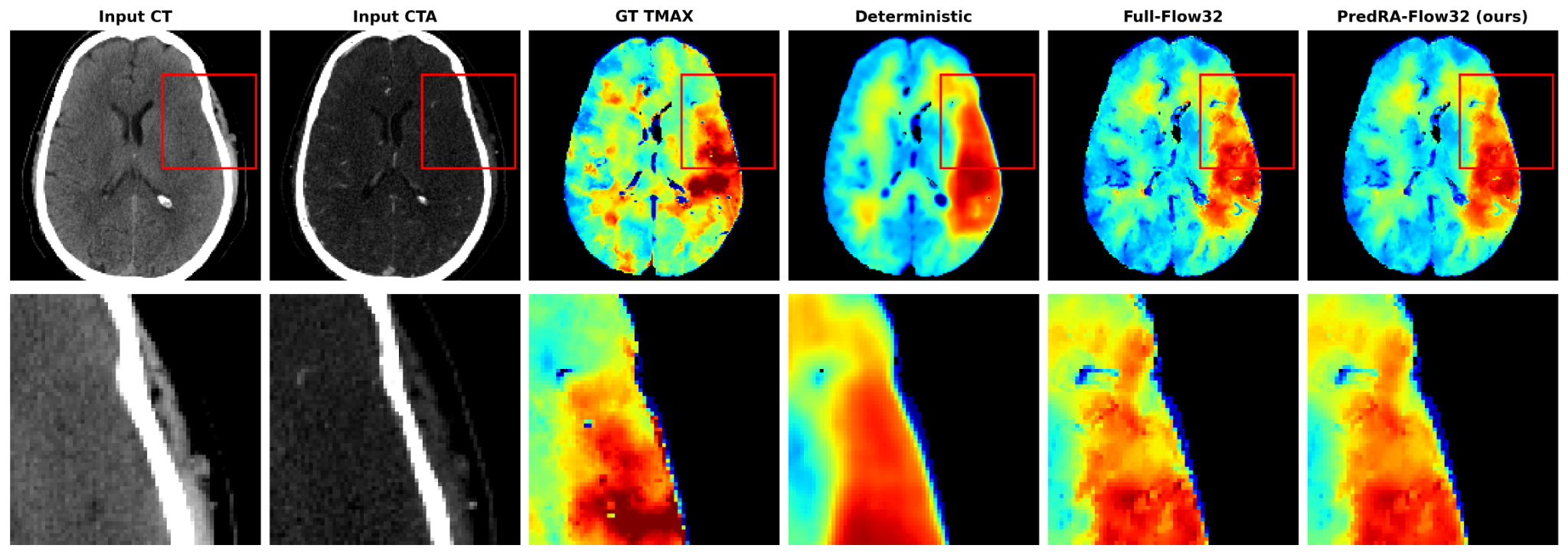}
\caption{Focused Tmax example comparing the deterministic anchor, the unallocated Full-Flow32-12.63M proposal, and PredRA-Flow32-12.63M. The first row shows the full CT and CTA inputs and the corresponding target and predictions. The second row uses the same square region for every panel. The deterministic prediction captures the broad perfusion pattern, Full-Flow32-12.63M adds finer local structure, and PredRA-Flow32-12.63M uses the same generated residual with limited amplitude.}
\label{fig:tmax_zoom}
\end{figure*}

\subsection{SynthRAD Deterministic and Residual-Backend Capacity Analyses}
The deterministic validation sweep shows a non-monotonic capacity response. DET-15.25M (15.254M parameters) gives the best validation SSIM and MAE among the tested deterministic widths, while the 29.893M and 60.997M models do not improve consistently. This justifies fixing DET-15.25M before residual-backend scaling.

\begin{table}[t]
\caption{Representative SynthRAD deterministic-capacity points on the full validation set. DET-15.25M is the validation-selected anchor. The complete seven-capacity sweep, including architecture width $b$, is reported in Appendix Table~\ref{tab:appendix_detcapacity_full}.}
\label{tab:detcapacity}
\centering
\tablefont
\setlength{\tabcolsep}{2.8pt}
\begin{tabular}{lrrrr}
\toprule
Model & MAE$\downarrow$ & SSIM$\uparrow$ & PSNR$\uparrow$ & LPIPS$\downarrow$\\
\midrule
DET-2.44M  & 0.1029 & 0.7385 & 17.7259 & 0.1933\\
DET-9.76M  & 0.0999 & 0.7446 & 17.9875 & 0.1847\\
\textbf{DET-15.25M} & \textbf{0.0967} & \textbf{0.7596} & \textbf{18.2548} & 0.1750\\
DET-61.00M & 0.0975 & 0.7537 & 18.1865 & \textbf{0.1750}\\
\bottomrule
\end{tabular}
\end{table}

With DET-15.25M fixed, both DDPM and Flow improve overall as residual-backend capacity increases from 1.547M to 12.628M, although the trend is not strictly monotonic; the 8.851M Flow point is a local regression. The representative endpoints are shown in Table~\ref{tab:backendcapacity}, with the complete sweep in Appendix Table~\ref{tab:appendix_backendcapacity_full}.

\begin{table}[t]
\caption{Representative post hoc SynthRAD frozen-test residual-backend capacity sensitivity with DET-15.25M fixed. Each $\lambda$ is selected independently on validation; residual capacity itself is not selected from these test curves. The complete DDPM/Flow capacity sweep is reported in Appendix Table~\ref{tab:appendix_backendcapacity_full}; full ($\lambda=1$) results are reported in Appendix Table~\ref{tab:appendix_fullbackendcapacity}.}
\label{tab:backendcapacity}
\centering
\tablefont
\setlength{\tabcolsep}{1.15pt}
\begin{tabular}{@{}lrrrrr@{}}
\toprule
Residual model & $\lambda$ & MAE$\downarrow$ & SSIM$\uparrow$ & PSNR$\uparrow$ & LPIPS$\downarrow$\\
\midrule
PredRA-DDPM-1.55M & 0.4000 & 0.1155 & 0.7436 & 17.2249 & 0.1618\\
PredRA-DDPM-12.63M & 0.4500 & 0.1147 & 0.7490 & 17.3000 & 0.1537\\
\midrule
PredRA-Flow32-1.55M & 0.4000 & 0.1151 & 0.7442 & 17.2541 & 0.1649\\
PredRA-Flow32-8.85M & 0.4500 & 0.1152 & 0.7451 & 17.2450 & 0.1599\\
\textbf{PredRA-Flow32-12.63M} & \textbf{0.5000} & \textbf{0.1144} & \textbf{0.7494} & \textbf{17.3293} & \textbf{0.1537}\\
\bottomrule
\end{tabular}
\end{table}

The corresponding full ($\lambda=1$) results reinforce the matched-capacity backend comparison. At 12.628M residual parameters, Full-Flow32-12.63M improves all four metrics relative to Full-DDPM-12.63M. This shows that the small advantage of allocated Flow at the 12.63M residual-capacity point is not produced only by different selected $\lambda$ values.

The complete full ($\lambda=1$) capacity sweep is moved to Appendix Table~\ref{tab:appendix_fullbackendcapacity} to keep the main results compact.

\begin{figure}[t]
\centering
\includegraphics[width=0.9800\columnwidth]{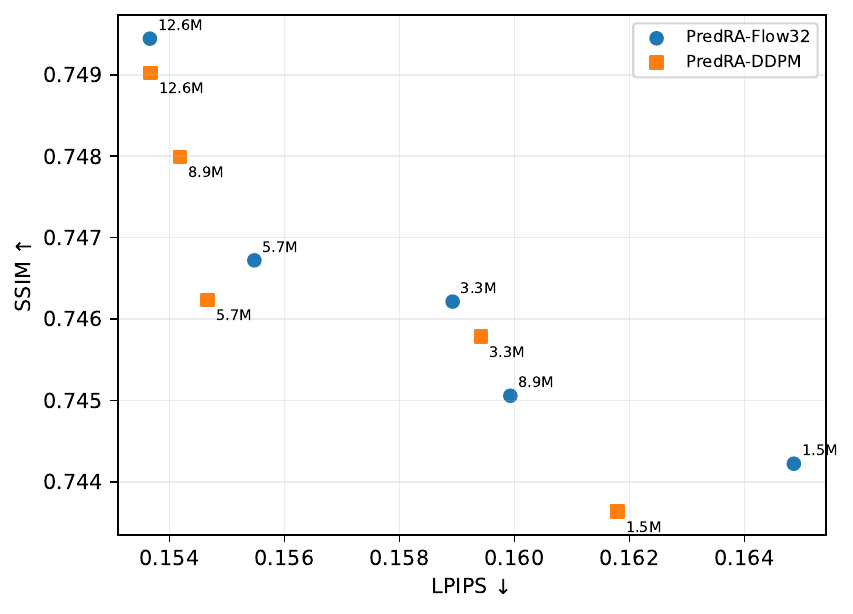}
\caption{SynthRAD residual-backend capacity sensitivity with DET-15.25M fixed. Points show the SSIM--LPIPS settings; labels give residual-network parameter counts. Markers are not connected because the response is non-monotonic. The 12.63M PredRA-Flow32 and PredRA-DDPM refinement points are nearly coincident.}
\label{fig:backendcapacity}
\end{figure}

\subsection{Flow Sampling-Step Analysis}
The 1.55M-backend Flow analysis exposes a clear quality--efficiency continuum. A one-step flow solution gives the strongest paired fidelity metrics, whereas increasing the number of integration steps progressively improves LPIPS while reducing MAE/SSIM/PSNR. Validation-selected residual refinement changes this trajectory. In this post hoc 1.55M sweep, 32 steps lies near a late-stage fidelity--perception trade-off point, while 64 steps yields only a small additional LPIPS gain at a measurable paired fidelity cost. The complete curve is reported as sensitivity analysis only and is not used to claim that 32 steps was selected by a frozen validation-only architecture search.

\begin{table*}[t]
\caption{SynthRAD Flow sampling-step sensitivity using the 1.55M residual backend and DET-15.25M. Full-Flow uses $\lambda=1$; \methodflow{} uses the independently validation-selected $\lambda$ shown for each step count.}
\label{tab:flowsteps}
\centering\tablefont
\setlength{\tabcolsep}{2.2pt}
\begin{tabular}{rcrrrrrrrrr}
\toprule
Steps & $\lambda$ & \multicolumn{4}{c}{Full-Flow} & \multicolumn{4}{c}{\methodflow{}}\\
\cmidrule(lr){3-6}\cmidrule(lr){7-10}
& & MAE$\downarrow$ & SSIM$\uparrow$ & PSNR$\uparrow$ & LPIPS$\downarrow$ & MAE$\downarrow$ & SSIM$\uparrow$ & PSNR$\uparrow$ & LPIPS$\downarrow$\\
\midrule
1  & 0.9500 & 0.1128 & 0.7571 & 17.5260 & 0.1858 & 0.1127 & 0.7572 & 17.5257 & 0.1858\\
2  & 0.5500 & 0.1146 & 0.7493 & 17.3139 & 0.1786 & 0.1134 & 0.7541 & 17.4293 & 0.1824\\
4  & 0.6500 & 0.1183 & 0.7338 & 16.9401 & 0.1693 & 0.1152 & 0.7461 & 17.2474 & 0.1740\\
8  & 0.5500 & 0.1217 & 0.7207 & 16.6486 & 0.1628 & 0.1157 & 0.7435 & 17.2100 & 0.1686\\
16 & 0.4500 & 0.1241 & 0.7109 & 16.4274 & 0.1595 & 0.1154 & 0.7440 & 17.2361 & 0.1664\\
32 & 0.4000 & 0.1254 & 0.7043 & 16.3045 & 0.1571 & 0.1151 & 0.7442 & 17.2541 & 0.1649\\
64 & 0.4000 & 0.1266 & 0.7003 & 16.2198 & 0.1568 & 0.1154 & 0.7432 & 17.2321 & 0.1630\\
\bottomrule
\end{tabular}
\end{table*}

\begin{table*}[t]
\caption{Standardized representative-forward efficiency analysis for the reported 12.63M PredRA configurations and external SynthRAD baselines. These PredRA configurations combine DET-15.25M with a 12.63M residual backend and contain 27.882M unique parameters; only the 12.628M residual network is optimized during residual-stage training because the deterministic anchor is frozen. The Flow32 realization is the fast PredRA operating variant, requiring 32 residual-network evaluations rather than the 1000 ancestral DDPM steps. Net evaluation time denotes one deterministic-anchor forward pass plus one native residual-network evaluation at $192\times192$, batch size 1; it is an per-network measure and does not represent complete sampling wall time. GFLOPs marked with $\geq$ are lower bounds when the profiler does not account for all operators.}
\label{tab:computecost}
\centering
\tablefont
\setlength{\tabcolsep}{4.2pt}
\begin{tabular}{lrrrrrr}
\toprule
Method & Params (M) & Trainable (M) & Sampler steps & GFLOPs/native eval & Net eval (ms) & Peak infer GPU (GB)\\
\midrule
PredRA-DDPM-12.63M & 27.8820 & 12.6280 & 1000 & $\geq$116.5470 & 17.9030 & 0.2710\\
PredRA-Flow32-12.63M & 27.8820 & 12.6280 & 32 & $\geq$116.5470 & 17.7570 & 0.2710\\
SelfRDB & 39.7270 & 39.7270 & -- & $\geq$71.2600 & 21.1320 & 1.8710\\
YODA & 53.5070 & 53.5070 & -- & $\geq$182.3400 & 13.8380 & 3.1360\\
I2I-Mamba & 109.3010 & 109.3010 & -- & $\geq$169.8630 & 9.3290 & 0.9620\\
ResViT & 123.4420 & 123.4420 & -- & $\geq$119.8000 & 12.6290 & 0.5170\\
MADM & 330.8140 & 330.8140 & -- & $\geq$308.7920 & 13.7980 & 1.2280\\
\bottomrule
\end{tabular}
\end{table*}

\subsection{Spatial Refinement Analysis}
Ground-truth-informed analyses on the primary CTP task indicated spatially heterogeneous residual utility, but test-time utility prediction remained weak. An early learned gate showed essentially no ranking association, and a later within-trajectory analysis reached only about $0.1100$ test Spearman correlation; neither setting beat global selected-strength control. The result is therefore retained as a negative mechanism experiment rather than as a component of the final controller.

\subsection{Qualitative Results}
Figure~\ref{fig:qualitative_multitask} provides fixed examples from all three translation settings under one layout. Deterministic outputs are generally smooth and faithful to source-conditioned structure, while full residual proposals and external generative baselines can add detail but may also change the paired image more aggressively. PredRA lies between these extremes because it keeps the deterministic reference and adds only the validation-supported amount of residual refinement.

Figure~\ref{fig:tmax_zoom} shows the mechanism more closely in one Tmax case. The deterministic prediction captures the large-scale perfusion pattern but is smoother than the target. Full-Flow32-12.63M adds stronger local structure. PredRA-Flow32-12.63M uses the same residual proposal but accepts only the amount supported by validation, keeping more paired structure while still recovering useful detail. The figure is illustrative; quantitative comparisons use the complete frozen cohorts.

\section{Discussion}
The main idea of PredRA is simple: deterministic prediction should not be thrown away when a generative model is introduced. The deterministic model already captures much of the source-supported target. What remains in the residual can still contain useful stable structure, but it can also contain sampling-dependent variation. PredRA therefore uses generation to recover the residual, then accepts only the amount of refinement that improves the final paired prediction.

The residual decomposition in Eq.~\eqref{eq:resdecomp} helps explain this behavior. If the deterministic model were perfect, the source-conditioned residual mean would vanish. In practice it does not, so $b_\theta(X)$ can contain additional predictable information. The generator is able to recover some of this information, but a full generative correction can also add unnecessary variation. This is why the best result can occur at an intermediate $\lambda$ rather than at either extreme.

The CTP results show this clearly. For both DDPM and Flow, PredRA improves MAE, SSIM, and PSNR over the matched full residual proposal, with only a small LPIPS change. No method is best on every metric: YODA is strongest on several distortion metrics, Full-Flow32-12.63M gives the lowest LPIPS overall, and PredRA-Flow32-12.63M gives a stronger balance between paired fidelity and perceptual quality. The contribution is therefore not a universal-best backbone, but a simple way to keep useful generated detail without accepting the full amount of generative variation.

The DWI+ADC task gives the same message on different targets. PredRA improves all four macro metrics for both matched DDPM and Flow backends. Because several external baseline schedules were shortened, the strongest evidence on this task is the matched Full-versus-PredRA comparison, where the residual network is unchanged and only the accepted refinement strength changes.

The SynthRAD experiments show that the idea is not tied to one generator. Both DDPM and conditional Flow benefit from controlled residual refinement. Flow32 also gives the fast PredRA variant. The capacity studies show that larger models alone do not solve the problem: deterministic performance eventually saturates, and larger residual models still benefit from controlling how much of their output is used.

The Flow step study adds another useful observation. The number of sampling steps changes the residual proposal itself, while $\lambda$ changes how much of one fixed proposal is accepted. These are different controls. More steps can improve perceptual similarity but may reduce paired fidelity; a smaller $\lambda$ can recover a better paired operating point from the same proposal. The relation $\mathcal{U}_{\mathrm{post}}(\lambda)\propto\lambda^2$ gives a simple mathematical explanation for why stronger residual scaling also admits more sampling variability.

The mechanism studies also explain why we use one global $\lambda$. Residual usefulness varies across space, but the tested source- and trajectory-based features do not predict that local usefulness well enough to improve the final output. Some interventions that appear useful on oracle intermediate states also lose their benefit during free-running generation. A global validation rule is therefore more reliable than a learned voxel-wise gate in the current experiments.

Several limitations remain. The decomposition into stable and remaining residual terms is an interpretation of the source-conditioned residual, not a calibrated uncertainty model. With one paired target per input, model error cannot be cleanly separated from true conditional variability. The DWI+ADC task is registered paired stroke data rather than an external-site cohort, and some external baseline schedules are compute-limited. SynthRAD CT$\rightarrow$MRI uses the paired cohort in the reverse direction. Some capacity and step sweeps were extended after test inspection and are therefore reported only as sensitivity analyses. The spatial oracle is diagnostic, $\lambda$ is selected from a finite validation grid, LPIPS is only one perceptual metric, and the efficiency table reports per-network evaluation time rather than full iterative sampling wall time.

\section{Conclusion}
PredRA starts from a deterministic prediction and keeps it as the stable reference. A residual generator then recovers information that the first model missed. Because the generated residual can contain both useful stable structure and unnecessary sampling-dependent variation, validation selects how much of that correction should be added.

Across CT+CTA$\rightarrow$CTP, CT+CTA$\rightarrow$DWI+ADC, and SynthRAD CT$\rightarrow$MRI, this controlled refinement improves paired fidelity over the matched full residual proposal while keeping much of its perceptual benefit. The same idea works with both DDPM and conditional Flow, and Flow provides the fast variant. More complex spatial and trajectory controllers did not reliably improve on the simple global rule.

The practical message is direct: predict the easy part first, use generation to recover what is still missing, and add only the amount of residual refinement that validation shows is useful. Future work should further determine where, rather than only how much, residual refinement is beneficial.

\appendices
\newpage
\section{Mathematical Details of Deterministic Anchoring and Controlled Refinement}
\label{app:mathdetails}
This appendix gives four short derivations that support the main idea. The proofs use squared Euclidean risk or pre-projection sampling variance and assume that the same residual proposal is scaled across candidate $\lambda$ values. The actual controller is still selected by validation SSIM under the LPIPS constraint in Eq.~\eqref{eq:constrainedlambda}; the proofs only explain why keeping a deterministic reference and using partial residual refinement is reasonable.

\subsection{Squared-Error Predictable Component}
Let
\begin{equation}
m(X)=\mathbb{E}[Y\mid X]
\end{equation}
be the conditional mean, and let $f(X)$ be any square-integrable deterministic predictor. Adding and subtracting $m(X)$ gives
\begin{align}
\mathbb{E}\|Y-f(X)\|_2^2
&=\mathbb{E}\|Y-m(X)\|_2^2
 +\mathbb{E}\|m(X)-f(X)\|_2^2 \nonumber\\
&\quad+2\mathbb{E}\!\left[\left\langle Y-m(X),m(X)-f(X)\right\rangle\right].
\label{eq:app_risk_expand}
\end{align}
The cross term vanishes by conditioning on $X$:
\begin{align}
&\mathbb{E}\!\left[\left\langle Y-m(X),m(X)-f(X)\right\rangle\right] \nonumber\\
&\quad=\mathbb{E}\!\left[\left\langle
\mathbb{E}[Y-m(X)\mid X],m(X)-f(X)
\right\rangle\right]=0.
\end{align}
Therefore
\begin{equation}
\boxed{
\mathbb{E}\|Y-f(X)\|_2^2
=
\mathbb{E}\|Y-m(X)\|_2^2
+
\mathbb{E}\|m(X)-f(X)\|_2^2
}
\label{eq:app_pythagorean}
\end{equation}
and the unrestricted Bayes-optimal deterministic predictor under squared error is $f^{*}(X)=m(X)$ almost surely. Its Bayes residual $r_{\mathrm{B}}=Y-m(X)$ obeys
\begin{equation}
\mathbb{E}[r_{\mathrm{B}}\mid X]=0,
\qquad
\operatorname{Cov}(r_{\mathrm{B}}\mid X)=\operatorname{Cov}(Y\mid X).
\end{equation}
Thus a conditional-mean predictor removes the predictable first moment but need not remove conditional variation. In the actual method, Eq.~\eqref{eq:opdet} uses a finite network class and mixed fidelity objectives, so this result is used only to motivate the deterministic--residual division of responsibility; it does not assert that the trained anchor equals the exact conditional mean or that the remaining residual is purely aleatoric.

\subsection{Residual Scaling Controls Admitted Stochastic Variability}
Let $Z$ be the residual generator's sampling variable and define the unprojected allocated output
\begin{equation}
\bar Y_{\lambda}=f(X)+\lambda\tilde r(X,Z).
\end{equation}
Conditioned on $X$, $f(X)$ is constant with respect to $Z$. Therefore
\begin{align}
\operatorname{Cov}_{Z}(\bar Y_{\lambda}\mid X)
&=
\operatorname{Cov}_{Z}
\!\left(f(X)+\lambda\tilde r(X,Z)\mid X\right)\\
&=
\lambda^2\operatorname{Cov}_{Z}
\!\left(\tilde r(X,Z)\mid X\right).
\label{eq:app_var_scale}
\end{align}
Taking the trace and expectation over $X$ yields
\begin{equation}
\boxed{
\mathcal{U}_{\mathrm{post}}(\lambda)
=
\lambda^2\mathcal{U}_{\mathrm{post}}(1)
}
\label{eq:app_upost_scale}
\end{equation}
for the pre-projection uncertainty measure in Eq.~\eqref{eq:upost}. Hence $\lambda=0$ removes residual sampling variability from the output, while $\lambda=1$ admits the full proposal variability. The exact quadratic relation need not survive the nonlinear output-range projection, but $\lambda$ remains the direct amplitude control applied before projection. This result justifies interpreting the refinement strength as post-generation stochastic-responsibility control without claiming calibrated predictive uncertainty.

\begin{table*}[t]
\caption{Complete deterministic-capacity validation sweep. The selected DET-15.25M anchor corresponds to width $b=80$.}
\label{tab:appendix_detcapacity_full}
\centering\tablefont
\setlength{\tabcolsep}{4.2pt}
\begin{tabular}{lrrrrrr}
\toprule
Model & Width $b$ & Params (M) & MAE$\downarrow$ & SSIM$\uparrow$ & PSNR$\uparrow$ & LPIPS$\downarrow$\\
\midrule
DET-2.44M  & 32  & 2.4430  & 0.1029 & 0.7385 & 17.7259 & 0.1933\\
DET-3.82M  & 40  & 3.8160  & 0.1028 & 0.7324 & 17.7003 & 0.1951\\
DET-5.49M  & 48  & 5.4940  & 0.1016 & 0.7340 & 17.7611 & 0.1938\\
DET-9.76M  & 64  & 9.7640  & 0.0999 & 0.7446 & 17.9875 & 0.1847\\
\textbf{DET-15.25M} & \textbf{80} & \textbf{15.2540} & \textbf{0.0967} & \textbf{0.7596} & \textbf{18.2548} & 0.1750\\
DET-29.89M & 112 & 29.8930 & 0.0998 & 0.7428 & 18.0064 & 0.1840\\
DET-61.00M & 160 & 60.9970 & 0.0975 & 0.7537 & 18.1865 & \textbf{0.1750}\\
\bottomrule
\end{tabular}
\end{table*}

\subsection{Why Partial Residual Refinement Can Be Optimal}
Fix an arbitrary deterministic anchor $f(X)$ and define its true paired residual
\begin{equation}
r^{*}=Y-f(X).
\end{equation}
Write one realized residual-generator proposal as
\begin{equation}
\tilde r=r^{*}+e,
\end{equation}
where $e$ is its proposal error. Temporarily ignoring the output-domain projection, the allocated prediction is
\begin{equation}
\hat Y_{\lambda}^{\mathrm{unproj}}=f(X)+\lambda\tilde r.
\end{equation}
Since $Y=f(X)+r^{*}$,
\begin{equation}
\hat Y_{\lambda}^{\mathrm{unproj}}-Y
=-(1-\lambda)r^{*}+\lambda e.
\end{equation}
Define
\begin{equation}
A=\mathbb{E}\|r^{*}\|_2^2,
\qquad
B=\mathbb{E}\|e\|_2^2,
\qquad
C=\mathbb{E}\langle r^{*},e\rangle .
\end{equation}
The expected squared error is then
\begin{equation}
\mathcal{R}(\lambda)
=(1-\lambda)^2A+\lambda^2B-2\lambda(1-\lambda)C.
\label{eq:app_lambda_risk}
\end{equation}
When $A+B+2C=\mathbb{E}\|\tilde r\|_2^2>0$, differentiating Eq.~\eqref{eq:app_lambda_risk} gives the unconstrained minimizer
\begin{equation}
\lambda_{\mathrm{unc}}^{*}
=
\frac{A+C}{A+B+2C}.
\label{eq:app_lambda_general}
\end{equation}
With the operational constraint $\lambda\in[0,1]$, the squared-error optimum is the interval projection
\begin{equation}
\lambda_{\mathrm{MSE}}^{*}
=
\Pi_{[0,1]}\!\left(
\frac{A+C}{A+B+2C}
\right).
\label{eq:app_lambda_clipped}
\end{equation}
A particularly transparent special case is an uncorrelated proposal error, $C=0$, for which
\begin{equation}
\boxed{
\lambda_{\mathrm{MSE}}^{*}
=
\frac{A}{A+B}
}
\label{eq:app_lambda_uncorr}
\end{equation}
provided $A+B>0$. Hence a perfect residual proposal ($B=0$) favors $\lambda=1$, whereas any nonzero uncorrelated proposal error ($B>0$) produces an interior shrinkage value whenever $A>0$. As $B/A$ grows, the optimal residual responsibility decreases toward zero. This establishes that full residual injection is not generally implied by the residual formulation itself.

Equation~\eqref{eq:app_lambda_uncorr} is not used to set $\lambda$ in the experiments. The real objective is not squared error alone, and the residual error need not be uncorrelated with $r^{*}$. PredRA therefore uses the held-out validation rule in Eq.~\eqref{eq:constrainedlambda}, which instantiates the fidelity-first constrained objective in Eq.~\eqref{eq:unified_objective} under the prespecified SSIM--LPIPS criterion.

\subsection{Effect of the Output-Domain Projection}
The normalized task-specific target range used in Eq.~\eqref{eq:lambda} is a closed convex box. For Euclidean projection onto any closed convex set $\mathcal{Y}$,
\begin{equation}
\|\Pi_{\mathcal{Y}}(a)-\Pi_{\mathcal{Y}}(b)\|_2
\leq
\|a-b\|_2.
\end{equation}
Because the ground-truth target satisfies $Y\in\mathcal{Y}$ and therefore $\Pi_{\mathcal{Y}}(Y)=Y$, setting $a=\hat Y_{\lambda}^{\mathrm{unproj}}$ and $b=Y$ yields
\begin{equation}
\boxed{
\|\hat Y_{\lambda}-Y\|_2
\leq
\|\hat Y_{\lambda}^{\mathrm{unproj}}-Y\|_2
}.
\label{eq:app_projection}
\end{equation}
Thus range projection cannot increase Euclidean error to an in-range target. This property is specific to Euclidean distance and is not claimed to imply monotonic improvement in SSIM or LPIPS.

\section{Complete SynthRAD Capacity Analyses}
\label{app:capacityfull}
The main text reports only representative capacity points. For reproducibility, the complete sweeps are provided here. The architecture width $b$ is the number of channels in the first U-Net feature stage. It is an architectural width, \emph{not} a parameter count, and identical $b$ values yield different total parameter counts for the deterministic and residual networks. Main-text names therefore use actual parameter counts.

\begin{table*}[t]
\caption{Complete post hoc frozen-test PredRA residual-backend capacity sensitivity with DET-15.25M fixed. Width $b$ is reported only to document the architecture; model names use actual residual parameter counts. Bold marks the best value in each metric across the displayed sweep.}
\label{tab:appendix_backendcapacity_full}
\centering\tablefont
\setlength{\tabcolsep}{2.8pt}
\begin{tabular}{llrrrrrr}
\toprule
Backend & Width $b$ & Params (M) & $\lambda$ & MAE$\downarrow$ & SSIM$\uparrow$ & PSNR$\uparrow$ & LPIPS$\downarrow$\\
\midrule
DDPM & 32 & 1.5470 & 0.4000 & 0.1155 & 0.7436 & 17.2249 & 0.1618\\
DDPM & 48 & 3.3100 & 0.4000 & 0.1152 & 0.7458 & 17.2515 & 0.1594\\
DDPM & 64 & 5.7450 & 0.4500 & 0.1152 & 0.7462 & 17.2513 & 0.1547\\
DDPM & 80 & 8.8510 & 0.4500 & 0.1150 & 0.7480 & 17.2709 & 0.1542\\
DDPM & 96 & 12.6280 & 0.4500 & 0.1147 & 0.7490 & 17.3000 & 0.1537\\
\midrule
Flow32 & 32 & 1.5470 & 0.4000 & 0.1151 & 0.7442 & 17.2541 & 0.1649\\
Flow32 & 48 & 3.3100 & 0.4500 & 0.1149 & 0.7462 & 17.2769 & 0.1589\\
Flow32 & 64 & 5.7450 & 0.5000 & 0.1149 & 0.7467 & 17.2749 & 0.1555\\
Flow32 & 80 & 8.8510 & 0.4500 & 0.1152 & 0.7451 & 17.2450 & 0.1599\\
Flow32 & 96 & 12.6280 & 0.5000 & \textbf{0.1144} & \textbf{0.7494} & \textbf{17.3293} & \textbf{0.1537}\\
\bottomrule
\end{tabular}
\end{table*}

\begin{table*}[t]
\caption{Complete post hoc frozen-test unallocated residual-proposal ($\lambda=1$) capacity sensitivity with DET-15.25M fixed. Bold marks the best value in each metric across the displayed sweep.}
\label{tab:appendix_fullbackendcapacity}
\centering\tablefont
\setlength{\tabcolsep}{3.2pt}
\begin{tabular}{llrrrrrr}
\toprule
Backend & Width $b$ & Params (M) & MAE$\downarrow$ & SSIM$\uparrow$ & PSNR$\uparrow$ & LPIPS$\downarrow$ & Steps\\
\midrule
DDPM & 32 & 1.5470 & 0.1273 & 0.7007 & 16.1777 & 0.1560 & 1000\\
DDPM & 48 & 3.3100 & 0.1270 & 0.7042 & 16.1971 & 0.1530 & 1000\\
DDPM & 64 & 5.7450 & 0.1259 & 0.7088 & 16.2848 & 0.1502 & 1000\\
DDPM & 80 & 8.8510 & 0.1255 & 0.7121 & 16.3108 & 0.1491 & 1000\\
DDPM & 96 & 12.6280 & 0.1248 & 0.7139 & 16.3686 & 0.1481 & 1000\\
\midrule
Flow32 & 32 & 1.5470 & 0.1254 & 0.7043 & 16.3045 & 0.1571 & 32\\
Flow32 & 48 & 3.3100 & 0.1241 & 0.7116 & 16.4145 & 0.1519 & 32\\
Flow32 & 64 & 5.7450 & 0.1233 & 0.7160 & 16.4888 & 0.1500 & 32\\
Flow32 & 80 & 8.8510 & 0.1249 & 0.7098 & 16.3513 & 0.1536 & 32\\
Flow32 & 96 & 12.6280 & \textbf{0.1221} & \textbf{0.7204} & \textbf{16.5824} & \textbf{0.1471} & 32\\
\bottomrule
\end{tabular}
\end{table*}

\begin{table*}[t]
\caption{Per-target CT+CTA$\rightarrow$CTP results from the shared-evaluator validated benchmark. Each cell reports MAE / SSIM / PSNR / LPIPS.}
\label{tab:ctppertarget}
\centering
\scriptsize
\setlength{\tabcolsep}{3.0pt}
\begin{tabular}{lccc}
\toprule
Method & CBF & CBV & Tmax\\
\midrule
DET-15.25M & 0.2045 / 0.8287 / 12.9148 / 0.1720 & 0.1792 / 0.8304 / 13.7866 / 0.1651 & 0.1395 / 0.8450 / 15.2059 / 0.1511\\
Full-DDPM-12.63M & 0.2087 / 0.8037 / 12.4641 / 0.1241 & 0.1934 / 0.8005 / 12.7943 / 0.1174 & 0.1676 / 0.8116 / 13.6962 / 0.1163\\
PredRA-DDPM-12.63M & 0.1986 / 0.8169 / 13.0624 / 0.1260 & 0.1808 / 0.8160 / 13.5926 / 0.1189 & 0.1483 / 0.8325 / 14.7830 / 0.1182\\
Full-Flow32-12.63M & 0.1980 / 0.8083 / 12.8494 / 0.1181 & 0.1691 / 0.8051 / 13.6533 / 0.1121 & 0.1490 / 0.8217 / 14.6028 / 0.1120\\
PredRA-Flow32-12.63M & 0.1945 / 0.8199 / 13.1944 / 0.1208 & 0.1662 / 0.8192 / 14.0914 / 0.1145 & 0.1431 / 0.8322 / 14.9864 / 0.1145\\
SelfRDB & 0.2112 / 0.8093 / 12.0095 / 0.1333 & 0.2011 / 0.8065 / 12.4451 / 0.1276 & 0.1662 / 0.8179 / 13.5263 / 0.1257\\
I2I-Mamba & 0.2028 / 0.8057 / 12.3064 / 0.1310 & 0.1824 / 0.8067 / 13.0221 / 0.1195 & 0.1812 / 0.8115 / 13.1451 / 0.1258\\
ResViT & 0.2044 / 0.8114 / 12.1940 / 0.1216 & 0.1904 / 0.8073 / 12.8750 / 0.1184 & 0.1675 / 0.8190 / 13.5491 / 0.1160\\
MADM & 0.1944 / 0.8292 / 13.3092 / 0.1464 & 0.1788 / 0.8283 / 13.8057 / 0.1422 & 0.1500 / 0.8396 / 14.5839 / 0.1392\\
YODA & 0.1750 / 0.8501 / 14.1802 / 0.1253 & 0.1546 / 0.8449 / 14.9055 / 0.1223 & 0.1329 / 0.8613 / 15.7855 / 0.1134\\
\bottomrule
\end{tabular}
\end{table*}

\section{DWI+ADC Per-Target Results}
\label{app:dwiadcpertarget}
Table~\ref{tab:dwiadcpertarget} reports the DWI and ADC values underlying the macro benchmark in Table~\ref{tab:dwiadcmacro}. Each cell lists MAE / SSIM / PSNR / LPIPS in that order using the same frozen evaluator and patient-level aggregation.

\begin{table*}[t]
\caption{Per-target CT+CTA$\rightarrow$DWI+ADC results from the shared-evaluator benchmark. Each cell reports MAE / SSIM / PSNR / LPIPS.}
\label{tab:dwiadcpertarget}
\centering
\scriptsize
\setlength{\tabcolsep}{4.0pt}
\begin{tabular}{lcc}
\toprule
Method & DWI & ADC\\
\midrule
DET-15.25M & 0.0924 / 0.8414 / 26.1227 / 0.1402 & 0.2389 / 0.7652 / 16.3926 / 0.1376\\
Full-DDPM-12.63M & 0.1003 / 0.8194 / 24.1156 / 0.1427 & 0.2900 / 0.7359 / 14.4691 / 0.1197\\
PredRA-DDPM-12.63M & 0.0912 / 0.8433 / 26.1071 / 0.1346 & 0.2354 / 0.7689 / 16.4847 / 0.1220\\
Full-Flow32-12.63M & 0.0901 / 0.8136 / 25.4219 / 0.1424 & 0.3306 / 0.6865 / 13.6535 / 0.1537\\
PredRA-Flow32-12.63M & 0.0897 / 0.8434 / 26.4032 / 0.1331 & 0.2387 / 0.7654 / 16.4208 / 0.1371\\
SelfRDB & 0.1697 / 0.7798 / 18.9103 / 0.2196 & 0.3574 / 0.6549 / 12.9138 / 0.2207\\
I2I-Mamba & 0.1832 / 0.7657 / 17.7879 / 0.2372 & 0.3152 / 0.6801 / 14.0023 / 0.2385\\
ResViT & 0.1864 / 0.7590 / 17.8575 / 0.2455 & 0.3212 / 0.6688 / 13.9286 / 0.2945\\
MADM & 0.3269 / 0.7003 / 19.0243 / 0.2768 & 0.3073 / 0.7453 / 14.9450 / 0.2367\\
YODA & 0.1026 / 0.8260 / 24.4105 / 0.1655 & 0.2215 / 0.7836 / 17.1230 / 0.1135\\
\bottomrule
\end{tabular}
\end{table*}

\section{Mechanism Analyses Beyond the Deterministic Anchor}
\label{app:mechanism}

\subsection{Oracle-to-Free Reverse-Trajectory Reversal}
Additional reverse-trajectory checks confirmed that target-consistent gains can disappear after autonomous evolution. At start timestep 200, a representative comparison changed from an oracle-state SSIM improvement of $+0.0001$ to a free-running change of $-0.0001$. The strongest measured interaction occurred for a start timestep of 600 at $t=297$, where the interaction term was $-0.0088$ despite a positive intermediate-state improvement of $+0.0038$. These additional cases reinforce the main-text conclusion that interventions must be judged on the final autonomous sample rather than inferred from oracle-like intermediate states.

\subsection{Spatial Residual Utility and the Oracle--Deployable Gap}
Development-stage ground-truth-informed analyses indicated that the usefulness of a realized residual varies across space. Because those oracle diagnostics use target information and predate the shared final evaluator, their image-quality values are not mixed with the final benchmark tables. The deployable conclusion is more stable: an early learned utility score had essentially no ranking association (Spearman $\rho=-0.0092$), and a later within-trajectory analysis achieved only approximately $+0.1850/+0.1070$ validation/test Spearman correlation with its best baseline feature set; adding trajectory features did not improve validation correlation. Neither learned spatial controller outperformed simple global residual-strength selection. The reported PredRA configurations therefore use validation-selected scalar residual strength (per target where applicable), while the spatial oracle is retained only as mechanism evidence that local utility can vary.

\section{CTP Per-Target Results}
\label{app:ctppertarget}
Table~\ref{tab:ctppertarget} reports the per-target values underlying the macro CTP benchmark in Table~\ref{tab:final}. Each cell lists MAE / SSIM / PSNR / LPIPS in that order, using the same frozen evaluator and patient-level aggregation as the macro results.

\clearpage
\balance
\bibliographystyle{IEEEtran}
\bibliography{PredRA_references}

@inproceedings{isola2017pix2pix,
  author    = {Phillip Isola and Jun-Yan Zhu and Tinghui Zhou and Alexei A. Efros},
  title     = {Image-to-Image Translation with Conditional Adversarial Networks},
  booktitle = {Proceedings of the IEEE Conference on Computer Vision and Pattern Recognition (CVPR)},
  pages     = {1125--1134},
  year      = {2017}
}

@inproceedings{zhu2017cyclegan,
  author    = {Jun-Yan Zhu and Taesung Park and Phillip Isola and Alexei A. Efros},
  title     = {Unpaired Image-to-Image Translation Using Cycle-Consistent Adversarial Networks},
  booktitle = {Proceedings of the IEEE International Conference on Computer Vision (ICCV)},
  pages     = {2223--2232},
  year      = {2017}
}

@article{armanious2020medgan,
  author={Karim Armanious and Chenming Jiang and Marc Fischer and Thomas K{\"u}stner and Tobias Hepp and Konstantin Nikolaou and Sergios Gatidis and Bin Yang},
  title={{MedGAN}: Medical Image Translation using {GANs}}, journal={Computerized Medical Imaging and Graphics}, volume={79}, pages={101684}, year={2020}, doi={10.1016/j.compmedimag.2019.101684}}

@article{dalmaz2022resvit,
  author={Onat Dalmaz and Mahmut Yurt and Tolga {\c{C}}ukur}, title={{ResViT}: Residual Vision Transformers for Multimodal Medical Image Synthesis}, journal={IEEE Transactions on Medical Imaging}, volume={41}, number={10}, pages={2598--2614}, year={2022}, doi={10.1109/TMI.2022.3167808}}

@article{zhang2025unified,
  author  = {Yue Zhang and Chengtao Peng and Qiuli Wang and Dan Song and Kaiyan Li and S. Kevin Zhou},
  title   = {Unified Multi-Modal Image Synthesis for Missing Modality Imputation},
  journal = {IEEE Transactions on Medical Imaging},
  volume  = {44},
  number  = {1},
  pages   = {4--18},
  year    = {2025},
  doi     = {10.1109/TMI.2024.3424785}
}

@article{atli2026i2imamba,
  author={Omer F. Atli and Bilal Kabas and Fuat Arslan and Arda C. Demirtas and Mahmut Yurt and Onat Dalmaz and Tolga {\c{C}}ukur}, title={{I2I-Mamba}: Multi-modal Medical Image Synthesis via Selective State Space Modeling}, journal={IEEE Transactions on Biomedical Engineering}, year={2026}, note={Early Access}, doi={10.1109/TBME.2026.3698056}}

@inproceedings{ho2020ddpm,
  author    = {Jonathan Ho and Ajay Jain and Pieter Abbeel},
  title     = {Denoising Diffusion Probabilistic Models},
  booktitle = {Advances in Neural Information Processing Systems (NeurIPS)},
  year      = {2020}
}

@inproceedings{lipman2023flow,
  author    = {Yaron Lipman and Ricky T. Q. Chen and Heli Ben-Hamu and Maximilian Nickel and Matthew Le},
  title     = {Flow Matching for Generative Modeling},
  booktitle = {International Conference on Learning Representations (ICLR)},
  year      = {2023}
}

@inproceedings{song2021sde,
  author    = {Yang Song and Jascha Sohl-Dickstein and Diederik P. Kingma and Abhishek Kumar and Stefano Ermon and Ben Poole},
  title     = {Score-Based Generative Modeling through Stochastic Differential Equations},
  booktitle = {International Conference on Learning Representations (ICLR)},
  year      = {2021}
}

@inproceedings{saharia2022palette,
  author={Chitwan Saharia and William Chan and Huiwen Chang and Chris A. Lee and Jonathan Ho and Tim Salimans and David J. Fleet and Mohammad Norouzi}, title={Palette: Image-to-Image Diffusion Models}, booktitle={ACM SIGGRAPH 2022 Conference Proceedings}, pages={15:1--15:10}, year={2022}, doi={10.1145/3528233.3530757}}

@article{kazerouni2023survey,
  author  = {Amirhossein Kazerouni and Ehsan Khodapanah Aghdam and Moein Heidari and Reza Azad and Mohsen Fayyaz and Ilker Hacihaliloglu and Dorit Merhof},
  title   = {Diffusion Models in Medical Imaging: A Comprehensive Survey},
  journal = {Medical Image Analysis},
  volume  = {88},
  pages   = {102846},
  year    = {2023},
  doi     = {10.1016/j.media.2023.102846}
}

@inproceedings{blau2018tradeoff,
  author={Yochai Blau and Tomer Michaeli}, title={The Perception-Distortion Tradeoff}, booktitle={Proceedings of the IEEE Conference on Computer Vision and Pattern Recognition (CVPR)}, pages={6228--6237}, year={2018}}

@article{rassmann2026regression,
  author={Sebastian Rassmann and David K{\"u}gler and Christian Ewert and Martin Reuter}, title={Regression Is All You Need for Medical Image Translation}, journal={IEEE Transactions on Medical Imaging}, volume={45}, number={5}, pages={2156--2172}, year={2026}, doi={10.1109/TMI.2025.3650412}}

@inproceedings{li2023bbdm,
  author={Bo Li and Kaitao Xue and Bin Liu and Yu-Kun Lai}, title={{BBDM}: Image-to-Image Translation with Brownian Bridge Diffusion Models}, booktitle={Proceedings of the IEEE/CVF Conference on Computer Vision and Pattern Recognition (CVPR)}, pages={1952--1961}, year={2023}, doi={10.1109/CVPR52729.2023.00194}}

@inproceedings{yue2023resshift,
  author    = {Zongsheng Yue and Jianyi Wang and Chen Change Loy},
  title     = {{ResShift}: Efficient Diffusion Model for Image Super-Resolution by Residual Shifting},
  booktitle = {Advances in Neural Information Processing Systems (NeurIPS)},
  year      = {2023}
}

@inproceedings{liu2024rddm,
  author={Jiawei Liu and Qiang Wang and Huijie Fan and Yinong Wang and Yandong Tang and Liangqiong Qu}, title={Residual Denoising Diffusion Models}, booktitle={Proceedings of the IEEE/CVF Conference on Computer Vision and Pattern Recognition (CVPR)}, pages={2773--2783}, year={2024}, doi={10.1109/CVPR52733.2024.00268}}

@article{zhou2024cmdm,
  author={Yinchi Zhou and Tianqi Chen and Jun Hou and Huidong Xie and Nicha C. Dvornek and S. Kevin Zhou and David L. Wilson and James S. Duncan and Chi Liu and Bo Zhou}, title={Cascaded Multi-Path Shortcut Diffusion Model for Medical Image Translation}, journal={Medical Image Analysis}, volume={98}, pages={103300}, year={2024}, doi={10.1016/j.media.2024.103300}}

@article{chen2025madm,
  author={Tianqi Chen and Jun Hou and Yinchi Zhou and Huidong Xie and Xiongchao Chen and Qiong Liu and Xueqi Guo and Menghua Xia and James S. Duncan and Chi Liu and Bo Zhou}, title={2.5D Multi-View Averaging Diffusion Model for 3D Medical Image Translation: Application to Low-Count {PET} Reconstruction With {CT}-Less Attenuation Correction}, journal={IEEE Transactions on Medical Imaging}, volume={44}, number={11}, pages={4239--4250}, year={2025}, doi={10.1109/TMI.2025.3570342}}

@article{arslan2025selfrdb,
  author={Fuat Arslan and Bilal Kabas and Onat Dalmaz and Muzaffer Ozbey and Tolga {\c{C}}ukur}, title={Self-Consistent Recursive Diffusion Bridge for Medical Image Translation}, journal={Medical Image Analysis}, volume={106}, pages={103747}, year={2025}, doi={10.1016/j.media.2025.103747}}

@article{li2025fddm,
  author={Yunxiang Li and Hua-Chieh Shao and Xiaoxue Qian and You Zhang}, title={{FDDM}: Unsupervised Medical Image Translation with a Frequency-Decoupled Diffusion Model}, journal={Machine Learning: Science and Technology}, volume={6}, number={2}, pages={025007}, year={2025}, doi={10.1088/2632-2153/adc656}}

@inproceedings{lin2026drdd,
  author    = {Ziyue Lin and Jiahe Hou and Hongyu Xia and Xinrui Xie and Feifei Wang and Yuyin Zhou and Wei Wang and Jiawei Liu and Liangqiong Qu},
  title     = {Decoupled Residual Denoising Diffusion Models for Unified and Data Efficient Image-to-Image Translation},
  booktitle = {Proceedings of the IEEE/CVF Conference on Computer Vision and Pattern Recognition (CVPR)},
  pages     = {35967--35977},
  year      = {2026}
}

@inproceedings{cai2024mrdpm,
  author    = {Yuxin Cai and Jianhai Zhang and Lei He and Aravind Ganesh and Wu Qiu},
  title     = {Masked Residual Diffusion Probabilistic Model with Regional Asymmetry Prior for Generating Perfusion Maps from Multi-phase {CTA}},
  booktitle = {Medical Image Computing and Computer-Assisted Intervention (MICCAI)},
  pages     = {270--280},
  year      = {2024},
  doi       = {10.1007/978-3-031-72069-7_26}
}

@article{xue2025bibbdm,
  author  = {Kaitao Xue and Bo Li and Ziyi Liu and Zhifen He and Bin Liu and Congxuan Zhang and Yu-Kun Lai},
  title   = {{BiBBDM}: Bidirectional Image Translation With Brownian Bridge Diffusion Models},
  journal = {IEEE Transactions on Pattern Analysis and Machine Intelligence},
  volume  = {47},
  number  = {11},
  pages   = {10546--10559},
  year    = {2025},
  doi     = {10.1109/TPAMI.2025.3597667}
}

@article{ozbey2023syndiff,
  author  = {Muzaffer Ozbey and Onat Dalmaz and Salman U. H. Dar and Hasan A. Bedel and Saban Ozturk and Alper Gungor and Tolga Cukur},
  title   = {Unsupervised Medical Image Translation with Adversarial Diffusion Models},
  journal = {IEEE Transactions on Medical Imaging},
  volume  = {42},
  number  = {12},
  pages   = {3524--3539},
  year    = {2023},
  doi     = {10.1109/TMI.2023.3290149}
}

@article{li2024fgdm,
  author={Yunxiang Li and Hua-Chieh Shao and Xiao Liang and Liyuan Chen and Ruiqi Li and Steve Jiang and Jing Wang and You Zhang}, title={Zero-Shot Medical Image Translation via Frequency-Guided Diffusion Models}, journal={IEEE Transactions on Medical Imaging}, volume={43}, number={3}, pages={980--993}, year={2024}, doi={10.1109/TMI.2023.3325703}}

@article{wang2024midiffusion,
  author={Zihao Wang and Yingyu Yang and Yuzhou Chen and Tingting Yuan and Maxime Sermesant and Herv{\'e} Delingette and Ona Wu}, title={Mutual Information Guided Diffusion for Zero-Shot Cross-Modality Medical Image Translation}, journal={IEEE Transactions on Medical Imaging}, volume={43}, number={8}, pages={2825--2838}, year={2024}, doi={10.1109/TMI.2024.3382043}}

@article{luo2024tgdm,
  author  = {Yimin Luo and Qinyu Yang and Ziyi Liu and Zenglin Shi and Weimin Huang and Guoyan Zheng and Jun Cheng},
  title   = {Target-Guided Diffusion Models for Unpaired Cross-Modality Medical Image Translation},
  journal = {IEEE Journal of Biomedical and Health Informatics},
  volume  = {28},
  number  = {7},
  pages   = {4062--4071},
  year    = {2024},
  doi     = {10.1109/JBHI.2024.3393870}
}

@article{meng2024m2dn,
  author={Xiangxi Meng and Kaicong Sun and Jun Xu and Xuming He and Dinggang Shen}, title={Multi-Modal Modality-Masked Diffusion Network for Brain {MRI} Synthesis With Random Modality Missing}, journal={IEEE Transactions on Medical Imaging}, volume={43}, number={7}, pages={2587--2598}, year={2024}, doi={10.1109/TMI.2024.3368664}}

@inproceedings{zhang2024trajres,
  author={Junyu Zhang and Daochang Liu and Eunbyung Park and Shichao Zhang and Chang Xu}, title={Residual Learning in Diffusion Models}, booktitle={Proceedings of the IEEE/CVF Conference on Computer Vision and Pattern Recognition (CVPR)}, pages={7289--7299}, year={2024}, doi={10.1109/CVPR52733.2024.00696}}

@article{luo2024mgdm,
  author={Yimin Luo and Qinyu Yang and Yuheng Fan and Haikun Qi and Menghan Xia}, title={Measurement Guidance in Diffusion Models: Insight from Medical Image Synthesis}, journal={IEEE Transactions on Pattern Analysis and Machine Intelligence}, volume={46}, number={12}, pages={7983--7997}, year={2024}, doi={10.1109/TPAMI.2024.3399098}}

@inproceedings{zhang2025upsr,
  author    = {Leheng Zhang and Weiyi You and Kexuan Shi and Shuhang Gu},
  title     = {Uncertainty-guided Perturbation for Image Super-Resolution Diffusion Model},
  booktitle = {Proceedings of the IEEE/CVF Conference on Computer Vision and Pattern Recognition (CVPR)},
  pages     = {17980--17989},
  year      = {2025}
}

@inproceedings{teneggi2023trust,
  author={Jacopo Teneggi and Matthew Tivnan and J. Webster Stayman and Jeremias Sulam}, title={How to Trust Your Diffusion Model: A Convex Optimization Approach to Conformal Risk Control}, booktitle={Proceedings of the 40th International Conference on Machine Learning (ICML)}, series={Proceedings of Machine Learning Research}, volume={202}, pages={33940--33960}, year={2023}}

@inproceedings{kendall2017uncertainty,
  author    = {Alex Kendall and Yarin Gal},
  title     = {What Uncertainties Do We Need in Bayesian Deep Learning for Computer Vision?},
  booktitle = {Advances in Neural Information Processing Systems (NeurIPS)},
  year      = {2017}
}

@article{wang2004ssim,
  author  = {Zhou Wang and Alan C. Bovik and Hamid R. Sheikh and Eero P. Simoncelli},
  title   = {Image Quality Assessment: From Error Visibility to Structural Similarity},
  journal = {IEEE Transactions on Image Processing},
  volume  = {13},
  number  = {4},
  pages   = {600--612},
  year    = {2004},
  doi     = {10.1109/TIP.2003.819861}
}

@inproceedings{zhang2018lpips,
  author    = {Richard Zhang and Phillip Isola and Alexei A. Efros and Eli Shechtman and Oliver Wang},
  title     = {The Unreasonable Effectiveness of Deep Features as a Perceptual Metric},
  booktitle = {Proceedings of the IEEE Conference on Computer Vision and Pattern Recognition (CVPR)},
  pages     = {586--595},
  year      = {2018},
  doi       = {10.1109/CVPR.2018.00068}
}

@inproceedings{nie2017context,
  author    = {Dong Nie and Roger Trullo and Jun Lian and Caroline Petitjean and Su Ruan and Qian Wang and Dinggang Shen},
  title     = {Medical Image Synthesis with Context-Aware Generative Adversarial Networks},
  booktitle = {Medical Image Computing and Computer-Assisted Intervention (MICCAI)},
  volume    = {10435},
  pages     = {417--425},
  year      = {2017},
  doi       = {10.1007/978-3-319-66179-7_48}
}

@article{chartsias2018multimodal,
  author  = {Agisilaos Chartsias and Thomas Joyce and Mario Valerio Giuffrida and Sotirios A. Tsaftaris},
  title   = {Multimodal {MR} Synthesis via Modality-Invariant Latent Representation},
  journal = {IEEE Transactions on Medical Imaging},
  volume  = {37},
  number  = {3},
  pages   = {803--814},
  year    = {2018},
  doi     = {10.1109/TMI.2017.2764326}
}

@article{dar2019multicontrast,
  author  = {Salman U. H. Dar and Mahmut Yurt and Levent Karacan and Aykut Erdem and Erkut Erdem and Tolga Cukur},
  title   = {Image Synthesis in Multi-Contrast {MRI} With Conditional Generative Adversarial Networks},
  journal = {IEEE Transactions on Medical Imaging},
  volume  = {38},
  number  = {10},
  pages   = {2375--2388},
  year    = {2019},
  doi     = {10.1109/TMI.2019.2901750}
}

@inproceedings{wolterink2017unpaired,
  author    = {Jelmer M. Wolterink and Anna M. Dinkla and Mark H. F. Savenije and Peter R. Seevinck and Cornelis A. T. van den Berg and Ivana Isgum},
  title     = {Deep {MR} to {CT} Synthesis Using Unpaired Data},
  booktitle = {Simulation and Synthesis in Medical Imaging (SASHIMI)},
  series    = {Lecture Notes in Computer Science},
  volume    = {10557},
  pages     = {14--23},
  year      = {2017},
  doi       = {10.1007/978-3-319-68127-6_2}
}

@inproceedings{zhang2018shape,
  author    = {Zizhao Zhang and Lin Yang and Yefeng Zheng},
  title     = {Translating and Segmenting Multimodal Medical Volumes With Cycle- and Shape-Consistency Generative Adversarial Network},
  booktitle = {Proceedings of the IEEE Conference on Computer Vision and Pattern Recognition (CVPR)},
  pages     = {9242--9251},
  year      = {2018},
  doi       = {10.1109/CVPR.2018.00963}
}

@article{huo2019synseg,
  author  = {Yuankai Huo and Zhoubing Xu and Hyeonsoo Moon and Shunxing Bao and Albert Assad and Tamara K. Moyo and Michael R. Savona and Richard G. Abramson and Bennett A. Landman},
  title   = {{SynSeg-Net}: Synthetic Segmentation Without Target Modality Ground Truth},
  journal = {IEEE Transactions on Medical Imaging},
  volume  = {38},
  number  = {4},
  pages   = {1016--1025},
  year    = {2019},
  doi     = {10.1109/TMI.2018.2876633}
}

@inproceedings{liu2017unit,
  author    = {Ming-Yu Liu and Thomas Breuel and Jan Kautz},
  title     = {Unsupervised Image-to-Image Translation Networks},
  booktitle = {Advances in Neural Information Processing Systems (NeurIPS)},
  volume    = {30},
  year      = {2017}
}

@inproceedings{huang2018munit,
  author    = {Xun Huang and Ming-Yu Liu and Serge Belongie and Jan Kautz},
  title     = {Multimodal Unsupervised Image-to-Image Translation},
  booktitle = {European Conference on Computer Vision (ECCV)},
  pages     = {172--189},
  year      = {2018},
  doi       = {10.1007/978-3-030-01219-9_11}
}

@article{khader2023medical,
  author={Firas Khader and Gustav M{\"u}ller-Franzes and Soroosh Tayebi Arasteh and Tianyu Han and Christoph Haarburger and Maximilian Schulze-Hagen and Philipp Schad and Sandy Engelhardt and Bettina Bae{\ss}ler and Sebastian Foersch and Johannes Stegmaier and Christiane Kuhl and Sven Nebelung and Jakob Nikolas Kather and Daniel Truhn}, title={Denoising Diffusion Probabilistic Models for 3D Medical Image Generation}, journal={Scientific Reports}, volume={13}, pages={7303}, year={2023}, doi={10.1038/s41598-023-34341-2}}

@inproceedings{pinaya2022brain,
  author={Walter H. L. Pinaya and Petru-Daniel Tudosiu and Jessica Dafflon and Pedro F. Da Costa and Virginia Fernandez and Parashkev Nachev and S{\'e}bastien Ourselin and M. Jorge Cardoso}, title={Brain Imaging Generation with Latent Diffusion Models}, booktitle={Deep Generative Models, DGM4MICCAI 2022}, series={Lecture Notes in Computer Science}, volume={13609}, pages={117--126}, year={2022}, doi={10.1007/978-3-031-18576-2_12}}

@article{mullerfranzes2023medfusion,
  author  = {Gustav M{\"u}ller-Franzes and Jan Moritz Niehues and Firas Khader and Soroosh Tayebi Arasteh and Christoph Haarburger and Christiane Kuhl and Tianci Wang and Tianyu Han and Teresa Nolte and Sven Nebelung and Jakob Nikolas Kather and Daniel Truhn},
  title   = {A Multimodal Comparison of Latent Denoising Diffusion Probabilistic Models and Generative Adversarial Networks for Medical Image Synthesis},
  journal = {Scientific Reports},
  volume  = {13},
  pages   = {12098},
  year    = {2023},
  doi     = {10.1038/s41598-023-39278-0}
}

\end{document}